\documentclass[journal]{IEEEtran}
\usepackage{bm}
\usepackage{amsfonts}
\usepackage{subfigure}
\usepackage{graphicx}
\usepackage{epstopdf}
\usepackage{float}

\usepackage{rotating}
\usepackage{algorithm,algorithmicx}
\usepackage[noend]{algpseudocode}
\usepackage{longtable}
\usepackage{multirow}
\usepackage{booktabs}
\usepackage{amsmath} 
\usepackage{bbm}
\usepackage{array}
\ifCLASSINFOpdf
\else
\fi
\begin{document}
	%
	\title{Multi-Level Graph Convolutional Network with Automatic Graph Learning for Hyperspectral Image Classification}

	\author{Sheng~Wan,
		Chen~Gong,~\IEEEmembership{Member,~IEEE},
		Shirui~Pan,~\IEEEmembership{Member,~IEEE},
		Jie~Yang,
		and~Jian~Yang,~\IEEEmembership{Member,~IEEE}

		\thanks{S. Wan, C. Gong, and J. Yang are with the PCA Lab, the Key Laboratory of Intelligent Perception and Systems for High-Dimensional Information of Ministry of Education, the Jiangsu Key Laboratory of Image and Video Understanding for Social Security, and the School of Computer Science and Engineering, Nanjing University of Science and Technology, Nanjing, 210094, P. R. China. (e-mail: wansheng315@hotmail.com; chen.gong@njust.edu.cn; csjyang@njust.edu.cn). (\emph{Corresponding~author:~Chen~Gong.})}
		\thanks{J. Yang is with Institute of Image Processing and Pattern Recognition, Shanghai Jiao Tong University, Shanghai 200240, China (e-mail: jieyang@sjtu.edu.cn).}%
		\thanks{S. Pan is with Faculty of Information Technology, Monash University, Clayton, VIC 3800, Australia (e-mail: shirui.pan@monash.edu).}
	}

	\markboth{}%
	{Shell \MakeLowercase{\textit{et al.}}: Bare Demo of IEEEtran.cls for IEEE Journals}

	\maketitle
	
	\begin{abstract}
		Nowadays, deep learning methods, especially the Graph Convolutional Network (GCN), have shown impressive performance in hyperspectral image (HSI) classification. However, the current GCN-based methods treat graph construction and image classification as two separate tasks, which often results in suboptimal performance. Another defect of these methods is that they mainly focus on modeling the local pairwise importance between graph nodes while lack the capability to capture the global contextual information of HSI. In this paper, we propose a Multi-level GCN with Automatic Graph Learning method (MGCN-AGL) for HSI classification, which can automatically learn the graph information at both local and global levels. By employing attention mechanism to characterize the importance among spatially neighboring regions, the most relevant information can be adaptively incorporated to make decisions, which helps encode the spatial context to form the graph information at local level. Moreover, we utilize multiple pathways for local-level graph convolution, in order to leverage the merits from the diverse spatial context of HSI and to enhance the expressive power of the generated representations. To reconstruct the global contextual relations, our MGCN-AGL encodes the long range dependencies among image regions based on the expressive representations that have been produced at local level. Then inference can be performed along the reconstructed graph edges connecting faraway regions. Finally, the multi-level information is adaptively fused to generate the network output. In this means, the graph learning and image classification can be integrated into a unified framework and benefit each other. Extensive experiments have been conducted on three real-world hyperspectral datasets, which are shown to outperform the state-of-the-art methods.

	\end{abstract}

	%
	\IEEEpeerreviewmaketitle

	\section{Introduction}
	%
	%
	%
	%
	
	\IEEEPARstart{O}{ver} the past decades, hyperspectral imaging has witnessed a great surge of interest for Earth observations owing to its capacity to detect subtle spectral information, which makes it possible to discriminate different geographic objects \cite{Akhtar2018Nonparametric, Zhong2014Jointly}. As a consequence, hyperspectral image (HSI) classification, which aims to categorize each image pixel into a certain meaningful class according to the image contents, has attracted a growing interest in real-world applications, such as military target detection, vegetation monitoring, and disaster prevention and control \cite{Fauvel2013Advances}. However, the similarity occurring in the spectral bands among different land covers makes the classification task challenging \cite{Gong2019ACNN}. Confronted with this circumstance, spatial context is incorporated to generate discriminative spectral-spatial features, such as morphological profiles \cite{Benediktsson2005Classification, Fauvel2008Spectral} and spatial-based filtering techniques \cite{Gao2017Learning, Shen2011Three}, which generally employ handcrafted features followed by a classifier with predefined hyperparameters, and thereby requiring massive experts’ experience \cite{Zhang2018Diverse}.
	
	Recently, deep learning methods, which act dynamically to generate more robust and expressive feature representations than the handcrafted ones, have demonstrated their potentials in modeling the spectral-spatial features of HSI \cite{LiS2019Deep, Mou2017Deep}, such as Recurrent Neural Network (RNN) and Deep Belief Networks (DBN). Especially, Convolutional Neural Network (CNN) has shown great promise and been widely applied to HSI classification tasks. For instance, in \cite{Zhang2018Diverse}, a diversity of discriminative appearance factors are incorporated into CNN, in order to encode the semantic context-aware representations for generating promising features. Zhong \emph{et} \emph{al.} \cite{Zhong2018Spectral} improved the representation ability of CNN via designing the residual blocks which are capable of learning discriminative features from the spectral signatures and spatial contexts of HSI. Although much progress has been made on developing CNN-based HSI classification methods, the effectiveness of CNN is still limited in some irregular regions, such as class boundaries \cite{WanShMultiscale2019}. To be concrete, CNN cannot perceive the geometric variations between different object regions, since its convolution kernel is designed to only perform in regular squared regions. Additionally, the weights of a certain convolution kernel are kept identical when convolving all HSI patches, which inevitably causes a great loss of information around class boundaries and thus decreasing the representative power of the generated features. Therefore, the convolution kernels with fixed shapes and weights that are used in CNN cannot well adapt to the irregular structures in HSI. 
	
	To ameliorate this issue, Graph Convolutional Network (GCN) \cite{Zhang2020Deep, Wu2020Survey} has been utilized for HSI classification. Different from CNN, GCN can operate on graph-structured data, including social network data and graph-based representations of molecules \cite{Hamilton2017Inductive, Duvenaud2015Convolutional, Wang2019Attributed}, and it is able to pass, transform, and aggregate feature information across the graph nodes. With this, GCN can be naturally applied to non-Euclidean data, by which the class boundaries of different regions can be flexibly preserved. For example, in \cite{Qin2019Spectral}, GCN with the classical structure has been applied to HSI classification and achieved satisfying results. More recently, Wan \emph{et} \emph{al.} \cite{WanShMultiscale2019} utilized a dynamic GCN to exploit multi-scale spectral-spatial information, which outperformed several CNN-based methods. 
	
	Nevertheless, there still exist some common defects in these early-stage GCN-based methods. Specifically, the graph information is not originally available in HSI, and the direct approach to obtain one is manually constructing the graph based on pairwise Euclidean distance in advance, which has also been adopted by \cite{WanShMultiscale2019} and \cite{Qin2019Spectral}. However, the Euclidean distance may not be optimal to reveal the relationships among graph data \cite{Li2018Adaptive}, and thus the constructed graph may be either noisy or have edges that do not correspond to label agreement, which will ultimately weaken the expressive capacity of the generated representations. Besides, the above-mentioned methods fail to incorporate the global context, since they mainly focus on encoding the pairwise importance among local regions while disregarding the long range dependencies.

	To address these problems and further boost the performance of GCN-based HSI classification, we propose the Multi-level Graph Convolutional Network with Automatic Graph Learning (MGCN-AGL) method, where local spatial importance and global contextual information among graph nodes can be automatically learned in a unified framework. To precisely exploit the relationships among local regions, the proposed model adaptively characterizes the pairwise importance with learnable scaling coefficients during training. Then graph information at local spatial level can be automatically learned by the network, which can reduce the negative effect of an inaccurate pre-computed graph. As such, the model is able to focus on the most relevant spatial information of each region to make decisions. In addition, graph convolution governed by different spatial levels is performed to comprehensively capture the contextual information at multiple spatial levels. As a result, the regions with diverse object appearances can be better represented, which helps enhance the expressive power of the generated feature representations.
	
	Despite the critical role of spatial context in HSI classification \cite{Plaza2009Recent}, it is insufficient to pull in only the information from local spatial level, since regions that are far away in the 2D space may belong to the identical land-cover class. From that point of view, there exists a gap to close in merging the contextual information at both local and global levels. In the proposed method, we aim at reconstructing the topological graph information, in order to allow proper incorporation of global contextual information. Specifically, we employ the feature representations, which have been learned at local spatial level, for graph reconstruction, with the expectation that the strong expressive power will contribute to accurate reconstruction of the global graph information. Here, we enable the local-level graph convolution and the graph reconstruction to simultaneously operate in a unified framework. In this means, the topological graph information can also be learned automatically by the network. Then the inference is performed on the learned graph by passing messages between regions along the edges connecting them, by which the faraway regions can be connected as well. Hence, the feature representations can be progressively updated by aggregating the global contextual information appropriately. Finally, the features generated at local and global-level are integrated, balanced by learning, to obtain comprehensive representations.

	It is noteworthy that our proposed model jointly optimizes the representation learning and graph reconstruction, to the mutual benefit of both components. Concretely, the reconstructed topological graph information can be refined with expressive local-level feature representations, which will in turn improve the global-level representations. Furthermore, these two components operate collaboratively in a goal-directed manner, so that they can best fit for the subsequent node classification task. Experimental results on three typical real-world datasets confirm the superiority of our proposed MGCN-AGL to existing state-of-the-art methods.

	\section{Related Work}
	\label{Relatedworks}
	
	In this section, we will review some representative works on deep-learning-based HSI classification and GCN, since they are closely related to this paper.
	
	\subsection{Deep-Learning-based Hyperspectral Image Classification}
	
	As a state-of-the-art technique, deep learning \cite{Lecun2015Deep} has attracted increasing attention for its application to conventional computer vision tasks \cite{Zhao2019Object}. One main advantage is that deep learning techniques can automatically learn effective feature representations for a problem domain, thereby avoiding the complicated hand-crafted feature engineering \cite{Yang2018Hyperspectral}. In recent years, deep learning methods have also revolutionized the field of HSI classification \cite{Audebert2019Deep}, such as Stacked Auto-Encoder (SAE), Deep Belief Network (DBN), Recurrent Neural Network (RNN), and Generative Adversarial Networks (GAN) \cite{Chen2014Deep, Chen2015Spectral, Rodriguez1999Recurrent, Zhu2018Generative}.	
	
	Particularly, CNN, which is a class of neural networks with fewer parameters than fully-connected networks under the same number of hidden units, has demonstrated its superior performance for HSI classification, where the 2D CNN architecture has been widely studied. For example, in \cite{Makantasis2015Deep}, principal component analysis is employed to project the hyperspectral data into a three-channel tensor before a standard 2D CNN is applied. Alternatively, in \cite{Slavkovikj2015Hyperspectral}, the spatial dimensions of original hyperspectral data are flattened to generate a 2D image which can then be used as the input of a traditional 2D CNN. However, these methods force the HSI into the multimedia computer vision framework, thereby wasting the specific properties of HSI. Besides, 1D+2D CNN is another effective architecture for HSI classification \cite{Mou2018Unsupervised, Lee2017Going}. For instance, Luo \emph{et} \emph{al.} \cite{Luo2018HSI} introduced a CNN that performs spatial-spectral convolutions in the first layer for dimensionality reduction and employed a traditional 2D CNN to form the deeper layers that performs as usual. In addition to the 1D and 2D architecture, 3D CNN is developed to further boost the performance of HSI classification, which is capable of learning to recognize more complex 3D patterns of reflectances with fewer parameters and layers than 2D+1D CNN. For instance, \cite{Li2017Spectral} proposed to directly handle the hyperspectral cube with 3D CNN which works on three dimensions simultaneously via 3D convolutions. Although CNN-based methods have achieved promising performance for HSI classification, they simply apply fixed convolution kernels to different image regions, which will inevitably cause information loss in complex situations, and thus leading to imperfect classification results.
	
	\subsection{Graph Convolutional Network}
	
	As one of the hottest topics in graph-based deep learning, GCN defines convolutions and readout operations on irregular graph-structured data \cite{Learning2019Pan}. The convolutions on graphs can be roughly divided into two groups, namely spectral convolutions which perform convolution by transforming node representations into spectral domain with graph Fourier transform or its extensions, and spatial convolutions which are based on neighborhood aggregation \cite{Zhang2020Deep}. Spectral CNN \cite{Bruna2014Spectral} is the pioneering work of spectral methods, which converts signals defined in the vertex domain into spectral domain by leveraging graph Fourier transform and defines the convolution kernel as a set of learnable coefficients related with Fourier bases. However, this approach is based on the eigen-decomposition of the Laplacian matrix, thereby resulting in high computational complexity on large-scale graphs. Subsequently, ChebyNet \cite{Defferrard2016Convolutional} considers the convolution kernel as a polynomial function of the diagonal matrix containing the eigenvalues of Laplacian matrix. Afterwards, Kipf and Welling \cite{Kipf2016Semi} proposed a localized first-order approximation to ChebyNet, which contributes to more efficient filtering operation than that in spectral CNN.
	
	Different from spectral methods, the spatial methods directly define convolution in the vertex domain, following the practice of CNN. Concretely, convolution for each node is defined as a weighted average function over its neighboring nodes, where the weighting function characterizes the impact exerting to the target node by its neighboring nodes \cite{Xu2019Graph}. For example, in GraphSAGE \cite{Hamilton2017Inductive}, the weighting function is defined as various aggregators over neighboring nodes. In graph attention network \cite{Velivckovic2017Graph}, the weighting function can be adaptively learned via self-attention mechanism. Besides, MoNet \cite{Monti2017Geometric} considers convolution as a weighted average of multiple weighting functions defined over neighboring nodes, which offers a general framework for designing spatial methods.
	
	Over the past few years, GCN has demonstrated its superior performance in dealing with graph-structured data and achieved great success in several fields, such as social network mining \cite{Lei2009Relational} and natural language processing \cite{Zhou2018Commonsense}. More recently, GCN has also been applied to HSI classification, namely \cite{Qin2019Spectral} and \cite{WanShMultiscale2019}. However, these methods simply construct the graph based on Euclidean distance, which is a completely manual approach and may not uncover the intrinsic relationships among graph nodes. Besides, they mainly focus on encoding the pairwise importance among local regions while disregarding the long range dependencies of HSI.
	
	\begin{figure*}[!t]
		\centering
		\centering
		\includegraphics[width=15cm]{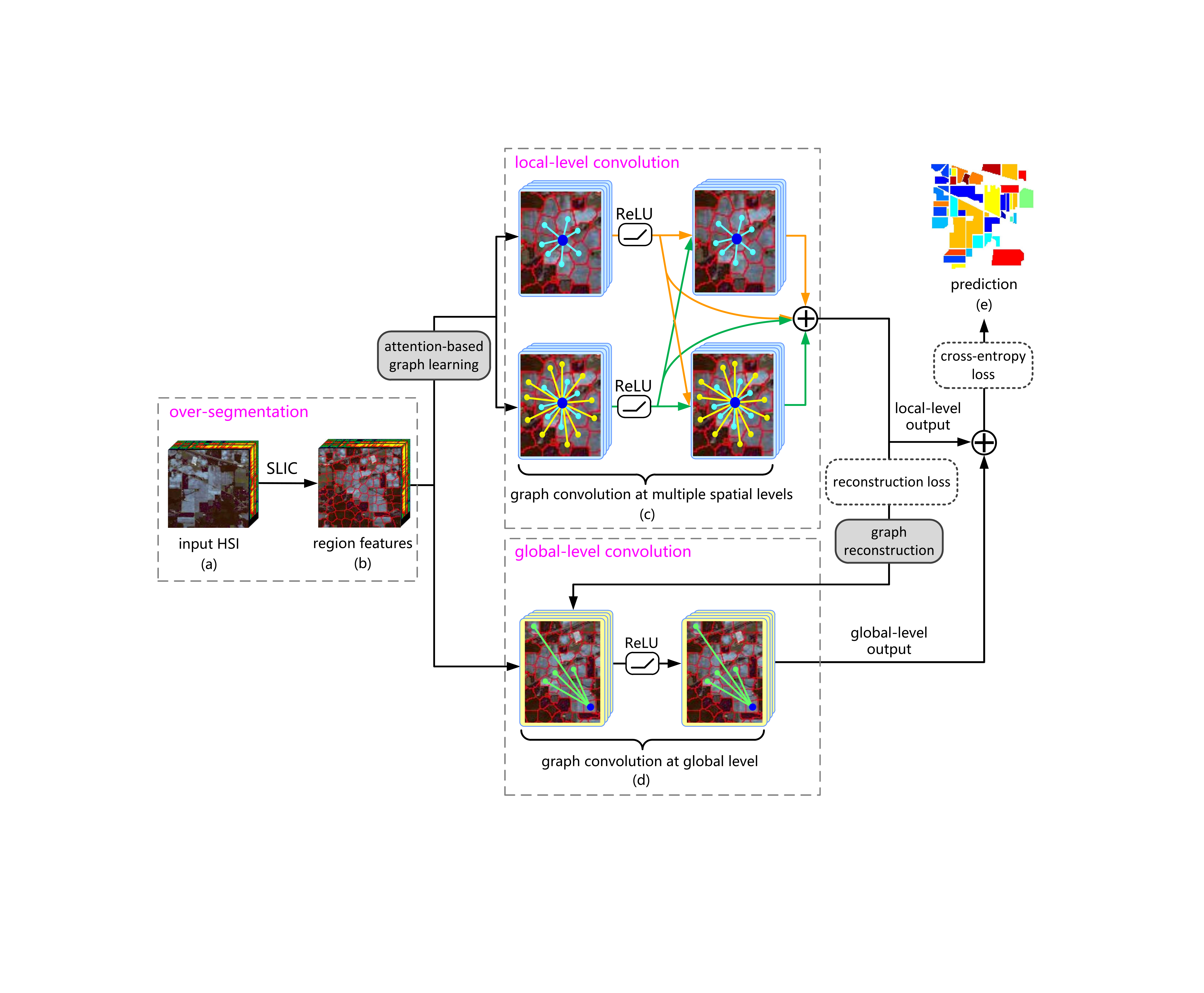}\hspace{0pt}
		\vskip -5pt
		\caption{The conceptual framework of our algorithm. (a) is the input hyperspectral data. (b) represents the regions obtained by over-segmenting the original HSI. (c) denotes the graph convolution at multiple spatial levels, where the pairwise importance among the regions can be learned with attention mechanism automatically. Here, ReLU \cite{Nair2010Rectified} is used as the activation function. (d) shows the global-level graph convolution, where the topological graph information is automatically reconstructed based on the representations generated at local level. In (e), the classification result is acquired by adaptively integrating the multi-level outputs.}
		\label{Overview}
	\end{figure*}

	\section{Proposed Method}
	\label{Proposedmethod}
	
	This section details our proposed MGCN-AGL algorithm, of which the schematic is exhibited in Fig.~\ref{Overview}. We first segment the input HSI (Fig.~\ref{Overview}(a)) into a set of compact regions (Fig.~\ref{Overview}(b)). Then graph convolution at multiple spatial levels (Fig.~\ref{Overview}(c)) is performed to obtain expressive feature representations. Subsequently, by reconstructing the topological graph information, we perform global-level convolution (Fig.~\ref{Overview}(d)) to capture long range dependencies among image regions. Finally, the classification result is produced by adaptively combining the outputs generated at different levels. In the following, the critical steps will be detailed by explaining the region-based segmentation technique (Section~\ref{sec_rbseg}), elaborating the automatic graph learning (Section~\ref{sec_convlo}), and describing the integration of multi-level contextual information (Section~\ref{sec_expglo}). Unless particularly specified, the important notations used in this paper are listed in Table~\ref{Nontations}.

	\begin{table}[!t]
		\centering
		\caption{Summary of the Notations}
		\begin{tabular}{cp{6.5cm}}
			\toprule
			\makebox[1.2cm][c] {\textbf{Notations}} & \makebox[6cm][c]{\textbf{Descriptions}} \\
			\midrule
			$\mathbf{X}$ & The feature matrix of all the graph nodes (image regions).\\
			$\mathbf{W}, \mathbf{a}$ & The learnable network parameters.\\
			$\mathbf{A}$ & The reconstructed graph adjacency matrix.\\
			$\mathbf{Z}_{b}^{(l)}$ & The representations generated from the $l^{\rm{th}}$ layer at branch $b$.\\
			$c_{ij}$ & The attention coefficient between node $\mathbf{x}_i$ and $\mathbf{x}_j$.\\

			\bottomrule
		\end{tabular}%
		\label{Nontations}%
	\end{table}%

	\subsection{Region-based Segmentation}
	\label{sec_rbseg}
	
	To construct a graph for HSI, determining each graph node with a image pixel is a common and simple approach. However, the efficiency of the subsequent graph convolution will be severely restricted due to the huge number of pixels. Consequently, in advance of classification, we employ a segmentation technique named SLIC \cite{Radhakrishna2012SLIC} to segment the original HSI into a set of compact homogeneous image regions, each of which consists of a small amount of pixels with strong spectral-spatial correlations. To be specific, the SLIC algorithm performs segmentation via iteratively growing the local clusters using a $k$-means algorithm. After the segmentation is completed, each image region is regarded as a graph node, and thus the number of graph nodes can be greatly reduced, which will accelerate the subsequent graph convolution. Here, the region features can be obtained by calculating the average spectral signatures of the involved pixels.

	\subsection{Automatic Graph Learning}
	\label{sec_convlo}
	
	As mentioned in the introduction, graph information is not originally available in hyperspectral data. A common approach to obtain a graph is calculating the pairwise Euclidean distance among the graph nodes (namely, image regions) in advance \cite{WanShMultiscale2019, Qin2019Spectral}. However, the existence of different types of noise in HSI may degrade the quality of the generated graph. Meanwhile, since the model training and graph construction are isolated steps, the obtained graph may not best fit the subsequent classification task. To ameliorate this issue, we propose to learn the graph information in an automatic manner from the network at both the local and global levels, which can be naturally integrated to the classification model.
	
\begin{figure}[!t]
		\centering
		\resizebox*{3.2cm}{5cm}{\includegraphics{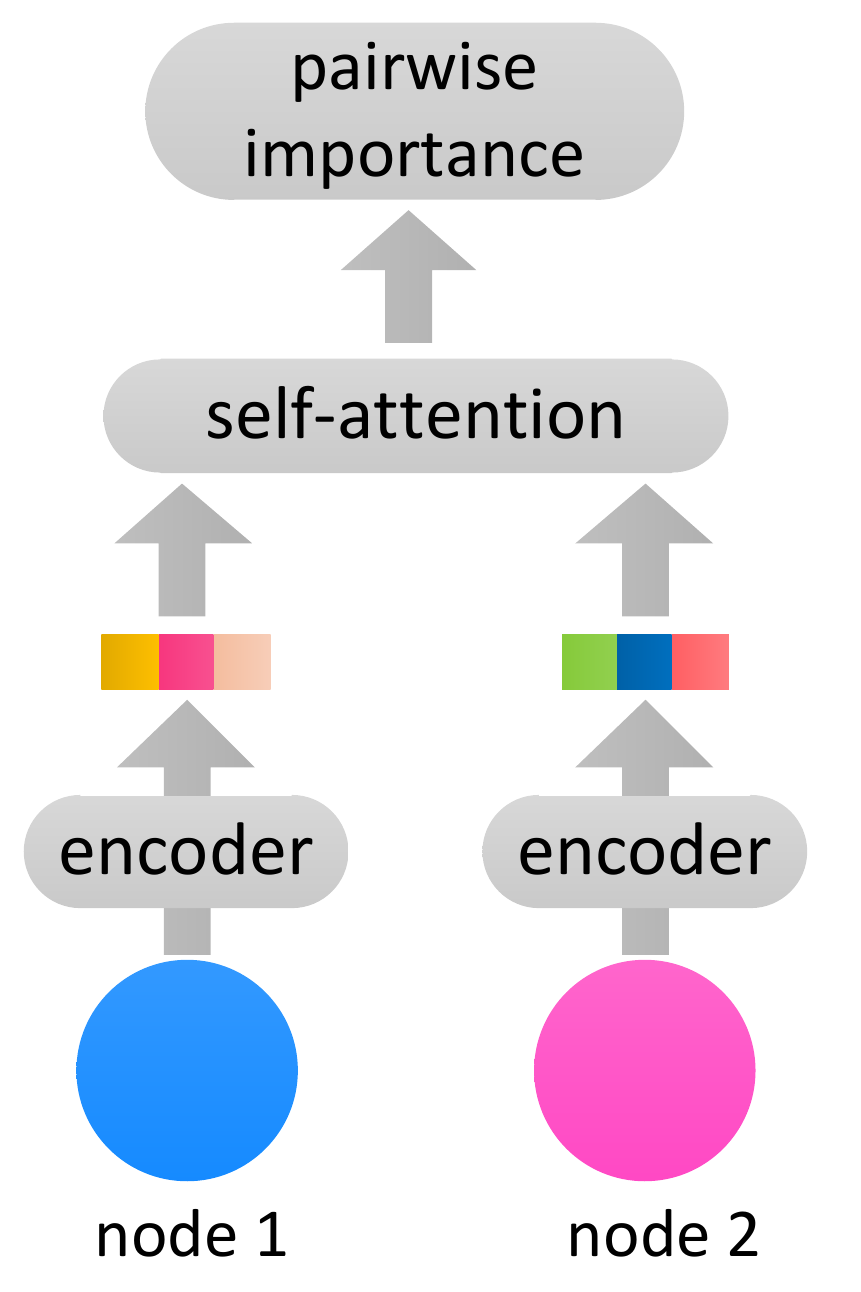}}\hspace{0pt}
		\caption{Graph information learning at local spatial level. Each of the two nodes acts as the other's spatial neighbor.}
		\label{fig_atten}
	\end{figure}

	To model the local spatial context of HSI, instead of using pre-computed fixed weight (e.g. Euclidean distance) as the measurement of pairwise importance, we resort to the attention mechanism to automatically capture the contextual relations among image regions. As an initial step, a shared linear transformation parametrized by a weight matrix $\mathbf{W}$ is applied to each node (i.e., region) $\mathbf{x}_i$ as an encoder, aiming at producing feature representations with sufficient expressive power. Then we perform self-attention on the encoded node features as follows:
	\begin{equation}
	\label{self_attention}
	{c_{ij}} = \mathbf{a}^{\top}[\mathbf{W}{\mathbf{x}_i}||\mathbf{W}{\mathbf{x}_j}],
	\end{equation}
	where the attention coefficient $c_{ij}$ reveals the pairwise importance between $\mathbf{x}_i$ and $\mathbf{x}_j$, $\mathbf{a}$ is a learnable weight vector, and $||$ denotes the operation to concatenate two vectors. In the most general formulation of Eq.~\eqref{self_attention}, each node is allowed to attend over all the other ones \cite{Velivckovic2017Graph}. In this paper, we inject the local spatial structure into the attention mechanism, in order to exploit the spatial context of HSI. In other words, we only compute the $c_{ij}$ for the regions $\mathbf{x}_j\in N(\mathbf{x}_i)$, with $N(\mathbf{x}_i)$ denoting the spatial neighborhood of $\mathbf{x}_i$. Then the attention coefficient $c_{ij}$ is normalized across all the spatial neighbors of $\mathbf{x}_i$ with a softmax function, namely
	\begin{equation}
	{\alpha _{ij}} = {\textstyle{{\exp ({c_{ij}})} \over {\sum\nolimits_{\mathbf{x}_k \in N({\mathbf{x}_i})} {\exp ({c_{ik}})} }}},
	\end{equation}
	in order to be easily comparable across different nodes. From the perspective of attribute values, the normalized attention coefficient $\alpha_{ij}$ can be represented as a single-layer feedforward neural network which is parametrized by a weight vector $\mathbf{a}$ with the LeakyReLU nonlinearity. Fully expanded out, the attention mechanism can be expressed as
	\begin{equation}
	\label{norm_alpha_ij}
	{\alpha _{ij}} = {\textstyle{{\exp ({\rm{LeakyReLU(}}{{\mathbf{a}}^ \top }{\rm{[\mathbf{W}}}{{\mathbf{x}}_i}{\rm{||\mathbf{W}}}{{\mathbf{x}}_j}{\rm{])}})} \over {\sum\nolimits_{\mathbf{x}_k \in N({\mathbf{x}_i})} {\exp ({\rm{LeakyReLU}}({{\mathbf{a}}^ \top }{\rm{[\mathbf{W}}}{{\mathbf{x}}_i}{\rm{||\mathbf{W}}}{{\mathbf{x}}_k}{\rm{]}}))} }}}.
	\end{equation}
	Fig.~\ref{fig_atten} explains how the pairwise importance between two nodes can be learned. Through utilizing the attention mechanism, our proposed model can automatically aggregate the important feature information from local spatial neighborhood. As a consequence, our graph learning model is less sensitive to the noise contained in the hyperspectral data, compared with \cite{Qin2019Spectral} and \cite{WanShMultiscale2019} which coarsely compute the pairwise similarities with fixed Euclidean distance.
	
\begin{figure}[!t]
		\centering
		\resizebox*{!}{6cm}{\includegraphics{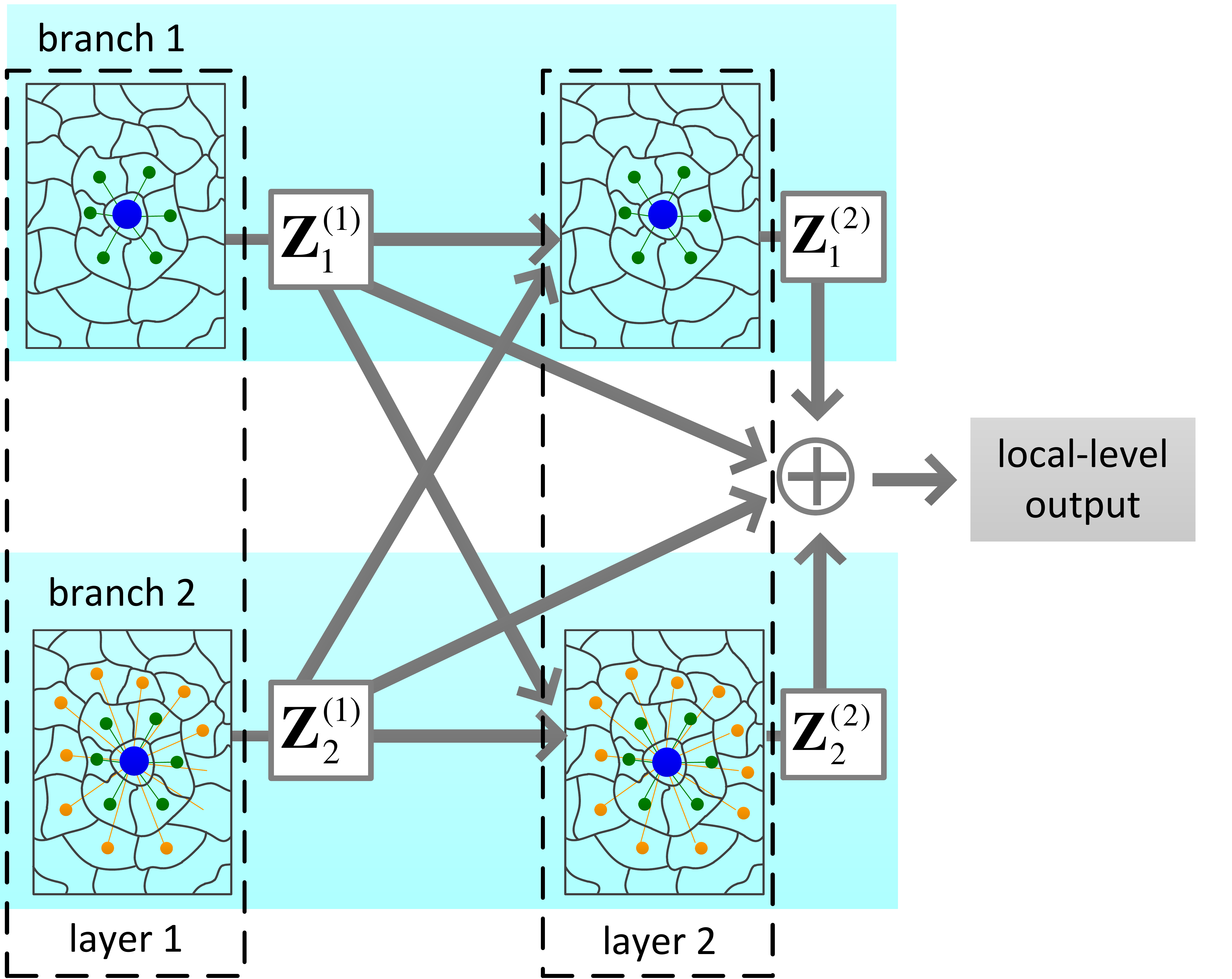}}\hspace{0pt}
		\caption{Exploitation of multi-level spatial information in our method. Graphs used in different branches comprise different neighborhood scales. $\mathbf{Z}_{1}^{(l)}$ and $\mathbf{Z}_{2}^{(l)}$ denote the representations generated from the ${l}^{\rm{th}}$ layer ($l=1\;\rm{or}\;2$) in branch 1 and 2, respectively. The green and orange nodes represent the spatial neighbors of the central blue one.}
		\label{fig_multi_spatial_lev}
	\end{figure}

	Nonetheless, there often exist various types of object appearances in HSI, where the object regions of the same land-cover class may even have diverse sizes and shapes \cite{Zhang2018Diverse}. For this reason, the above-mentioned graph learning model which incorporates contextual information from only a single spatial level is insufficient to obtain promising results. To cope with this issue, we intend to leverage the merits from multiple spatial levels, which is illustrated in Fig.~\ref{fig_multi_spatial_lev}, in order to better represent the image regions. In branch 1, the green nodes constitutes the 1-hop spatial neighbors of the central one. Meanwhile, the orange nodes together with the green nodes form the 2-hop spatial neighbors of the central one, as shown in branch 2. Here, we allow the representation $\mathbf{Z}_{1}^{(1)}$, which is generated from the ${1}^{\rm{st}}$ layer in branch 1, to propagate along three pathways. Under this circumstance, when $\mathbf{Z}_{1}^{(1)}$ propagates to branch 1, branch 2, or directly to the output, contextual information covering the 2-hop, 3-hop, or 1-hop spatial neighbors can be incorporated with the successive graph convolution. Analogously, $\mathbf{Z}_{2}^{(1)}$ is allowed to propagate along different pathways to involve the contextual information from different spatial levels. Note that the neighborhood sizes employed at branch 1 and branch 2 (namely, $s_1$ and $s_2$) can be adjusted for different HSIs, which will be discussed in Section~\ref{para_sen}. Then the overall steps of our local-level graph convolution can be acquired as
	\begin{equation}
	\label{Z_b_1}
	{\left[ {\mathbf{Z}_b^{(1)}} \right]_{i,:}} = \sigma \left( {\sum\limits_{\mathbf{x}_j \in {N_b}({\mathbf{x}_i})} {{\alpha _{ij}}\mathbf{W}_b^{({\rm{1}})}} {\mathbf{x}_j}} \right),
	\end{equation}
	\begin{equation}
	{\left[ {\mathbf{Z}_b^{(2)}} \right]_{i,:}} = \sigma \left( {\sum\limits_{\mathbf{x}_j \in {N_b}({\mathbf{x}_i})} {{\alpha _{ij}}\mathbf{W}_b^{(2)}} {{\left[ {{\mathbf{Z}^{(1)}}} \right]}_{j,:}}} \right),
	\end{equation}
	\begin{equation}
	\label{add_local_Z}
	{\mathbf{Z}_{loc}} = {\mathbf{Z}^{(1)}} + {\mathbf{Z}^{(2)}},
	\end{equation}
	where the branch index $b=1\;\rm{or}\;2$, $\mathbf{Z}_{loc}$ represents the local-level output, ${\mathbf{Z}^{(l)}} = \mathbf{Z}_1^{(l)} + \mathbf{Z}_2^{(l)}$ with the layer index $l=1\;\rm{or}\;2$, and $\sigma ( \cdot )$ is an activation function (e.g. the ReLU function \cite{Nair2010Rectified}). Besides, in branch $b$,  ${N_b}({\mathbf{x}_i})$ denotes the spatial neighborhood of $\mathbf{x}_i$ and $\mathbf{W}_b^{(l)}$ is the learnable weight matrix used in the ${l}^{\rm{th}}$ layer. With the automatic graph learning at multiple spatial levels, the expressive power of the obtained graph representations can be greatly enhanced, and thus the object regions with diverse shapes and sizes can be well represented.
	
	Nevertheless, the local convolution module fails to exploit the long range dependencies among image regions, thereby lacking the capacity to model the global context of HSI. Hence, we intend to explore the contextual relations beyond local level by reconstructing the topological graph information. To be specific, with the learned feature representations $\mathbf{Z}_{loc}$, the reconstructed graph adjacency matrix can be obtained as
	\begin{equation}
	\label{global_dense_A}
	{\widetilde{\mathbf{A}}_{ij}} = \exp \left({ - {\left\| {{{\left[ {{\mathbf{Z}_{loc}}} \right]}_{i,:}} - {{\left[ {{\mathbf{Z}_{loc}}} \right]}_{j,:}}} \right\|^2}}\right),
	\end{equation}
	where $\widetilde{\mathbf{A}}$ is able to encode the contextual relations between all region pairs, by which the long range dependencies among faraway regions can be captured. As such, the global-level contextual relations can be automatically learned by the network, and the expressive power of $\mathbf{Z}_{loc}$ helps to enhance the reliability of the reconstructed pairwise importance. Furthermore, we propose a reconstruction loss to improve the discriminative power of the graph, namely
	\begin{equation}
	\label{rec_err}
	{\mathcal{L}_r} = \sum\limits_{i,j \in {\mathbf{y}_G}} \ell{{{\left( {{{\widetilde {\mathbf{A}}}_{ij}} - {\mathbbm{1}_{[{y_i} = {y_j}]}}} \right)}}},
	\end{equation}
	where ${\mathbf{y}_G}$ denotes the set of indices corresponding to the labeled examples, $y_i$ represents the class label of $\mathbf{x}_i$, $\mathbbm{1}_{[{y_i} = {y_j}]}$ is an indicator function evaluating to 1 if ${y_i} = {y_j}$ and 0 otherwise, and $\ell$ is the squared error.
	
	In practice, a densely connected graph often leads to degraded classification performance, since the nodes that do not correspond to class agreement are connected by mistake. To address this issue, we only retain the graph edges with strong importance and remove the others, which can be expressed as
	\begin{equation}
	\label{global_sparse_A}
	{\mathbf{A}_{ij}} = \left\{ {\begin{array}{*{20}{c}}
		{{{\widetilde{\mathbf{A}}}_{ij}}} \,\;\;\;\;{\rm{if}}\;\;\,\;\widetilde{\mathbf{A}}_{ij}\ge\beta\\
		0 \;\;\;\;\,\;\;\;\;\rm{otherwise}\;\;\;\;\;
		\end{array}} \right.,
	\end{equation}
	where the parameter $\beta$ is fixed to 0.75 throughout the experiments. Then graph convolution at global level can be performed as
	\begin{equation}
	\label{Z_global_l}
	\mathbf{Z}_{glo}^{(l)} = \sigma \left( {\mathbf{A}\mathbf{X}\mathbf{W}_{glo}^{(l)}} \right),
	\end{equation}
	where $\mathbf{X}_{i,:}=\mathbf{x}_i$, $\mathbf{W}_{glo}^{(l)}$ is the learnable weight matrix used in the ${l}^{\rm{th}}$ graph convolutional layer, and $\mathbf{Z}_{glo}^{(l)}$ denotes the representations generated from the ${l}^{\rm{th}}$ layer. Since two graph convolutional layers have been utilized, the global-level output $\mathbf{Z}_{glo}$ can be acquired as
	\begin{equation}
	\label{Z_global}
	{\mathbf{Z}_{glo}} = \mathbf{Z}_{glo}^{(2)}.
	\end{equation}

	\subsection{Integration of Multi-Level Contextual Information}
	\label{sec_expglo}
	
	As is shown in Eq.~\eqref{add_local_Z}, the representations at multiple spatial levels are treated equally to calculate the local-level output $\mathbf{Z}_{loc}$, which inevitably neglects their different capacities in perceiving the variations of object appearances. To address this deficiency, Eq.~\eqref{add_local_Z} is modified to
	\begin{equation}
	\label{Z_loc}
	{{\mathbf{Z}}_{loc}} = \sum\limits_{l = 1}^2 {\sum\limits_{b = 1}^2 {\lambda_{b}^{(l)}}{\mathbf{Z}_{{b}}^{(l)}} }
	\end{equation}
	by assigning a parameter $\lambda_{b}^{(l)}$ to each $\mathbf{Z}_{b}^{(l)}$, where $\lambda_{b}^{(l)}$ can be learned via gradient descent. In this means, the representations generated at different spatial levels can adaptively contribute to the local-level output. Then a linear transformation parametrized by a weight matrix $\widehat{\mathbf{W}}$ is applied as follows:
	\begin{equation}
	{{\widehat{\mathbf{Z}}}_{loc}} = {{\mathbf{Z}}_{loc}}\widehat{\mathbf{W}},
	\end{equation}
	in order to make the feature dimension of the output representations consistent with the number of classes. Finally, the prediction of our model can be computed as
	\begin{equation}
	\label{O}
	\mathbf{O} = {\lambda _{loc}}{{\widehat{\mathbf{Z}}}_{loc}} + {\lambda _{glo}}{\mathbf{Z}_{glo}},
	\end{equation}
	where the parameters $\lambda_{loc}$ and $\lambda_{glo}$ are utilized to learn the importance of the contextual information at different levels.
	
	Apart from optimizing the reconstruction error of Eq.~\eqref{rec_err}, the cross-entropy error is adopted to penalize the differences between the network output and the labels of the originally labeled regions, namely
	\begin{equation}
	{\mathcal{L}_c} =  - \sum\limits_{i \in {{\mathbf{y}_G}}} {\sum\limits_{j = 1}^C {{\mathbf{Y}_{ij}}\ln {\mathbf{O}_{ij}}} }, 
	\end{equation}
	where $C$ is the number of classes, and $\mathbf{Y}$ represents the label matrix. Here, we let $\mathbf{Y}_{ij}=1$ if $\mathbf{x}_i$ belongs to the $j^{\rm{th}}$ class, and 0 otherwise. Then the overall loss function can be expressed as
	\begin{equation}
	\label{overall_error}
	\mathcal{L} = {\mathcal{L}_r} + \zeta {\mathcal{L}_c},
	\end{equation}
	where $\zeta$ is the coefficient assigned to the cross-entropy error and can be learned via gradient descent. In our proposed method, all the network parameters are updated through full-batch gradient descent \cite{WanShMultiscale2019}, and the implementation details are summarized in Algorithm~\ref{Algorithm1}.
	
	\begin{algorithm}[!t] 
		\caption{The Proposed MGCN-AGL for HSI Classification} 
		\label{Algorithm1} 
		\begin{algorithmic}[1] 
			\Require 
			Input image; number of iterations $\mathcal{T}$; learning rate $\eta$; the neighborhood sizes $s_1$ and $s_2$;
			\State Segment the whole image into superpixels via SLIC algorithm;
			\State // Train the MGCN-AGL model
			\For {$t=1$ to $\mathcal{T}$}
			\State {\bf{// Graph convolution at local level}}
			\State Perform graph learning at local spatial level by Eq.~\eqref{norm_alpha_ij};
			\State Calculate the local-level output $\mathbf{Z}_{loc}$ through Eq.~\eqref{Z_loc};
			\State {\bf{// Graph convolution at global level}}
			\State Reconstruct the global-level contextual relations based on $\mathbf{Z}_{loc}$ via Eqs.~\eqref{global_dense_A} and \eqref{global_sparse_A};
			\State Generate the output $\mathbf{Z}_{glo}$ by global-level convolution via Eqs.~\eqref{Z_global_l} and \eqref{Z_global};
			\State Integrate the multi-level information via Eq.~\eqref{O};
			\State Calculate the error terms according to Eq.~\eqref{overall_error}, and update the network parameters using full-batch gradient descent;
			\EndFor 
			\State \textbf{end for}
			\State Conduct label prediction based on the trained network;		
			\Ensure 
			Predicted label for each image region.
		\end{algorithmic} 
	\end{algorithm}	
	
	\section{Experimental Results}
	\label{Experiments}
	
	In this section, exhaustive experiments will be conducted to prove the effectiveness of the proposed method, and the corresponding analyses will also be provided. First, we compare MGCN-AGL with other state-of-the-art approaches on three real-world HSI datasets, where four metrics including per-class accuracy, Overall Accuracy (OA), Average Accuracy (AA), and kappa coefficient are used for performance evaluation. Afterwards, we investigate the impact of the number of labeled examples on OA. Then, the impact of the spatial neighborhood sizes on OA is analyzed. Finally, we demonstrate that the long range dependencies are advantageous for the model to improve the classification result. 
	
	\subsection{Datasets}
	\label{dataset}	
	
	The performance of our proposed MGCN-AGL is evaluated on three real-world benchmark datasets, i.e., the Houston University, the Indian Pines, and the Salinas, which will be introduced below.
	
	\subsubsection{Houston University}
	
	The Houston University dataset was acquired by the NSF-funded Center for Airborne Laser Mapping over the Houston University campus and its neighboring areas in 2012. This dataset was first known and distributed in the 2013 IEEE Geoscience and Remote Sensing Society Data Fusion Contest \cite{9046246}. It contains $349\times 1905$ pixels at a spatial resolution of 2.5 m and 144 spectral bands in the range of 380-1050 nm. There are 15 classes of interest, where the numbers of examples corresponding to each class are 1344, 1424, 730, 1234, 1268, 295, 1446, 1324, 1524, 1394, 1483, 1399, 540, 451, and 728, respectively.
	
	\subsubsection{Indian Pines}
	
	The second dataset utilized in the experiment is the well-known Indian Pines scene, which was gathered by AVIRIS sensor in 1992 and records north-western India. This dataset consists of $145\times145$ pixels with a spatial resolution of 20 m $\times$ 20 m, and there are 220 spectral channels with wavelength varying from 0.4 $\mu$m to 2.5 $\mu$m. As a usual step, 20 water absorption and noisy bands are removed, and the remaining 200 bands are retained. This dataset is challenging for traditional HSI classification methods due to the existence of highly mixed examples \cite{9046246}. The original ground truth of the Indian Pines dataset includes 16 land-cover classes, such as `Alfalfa', `Corn-notill', `Corn-mintill', etc. The numbers of examples in each land-cover class are 16, 1398, 800, 207, 453, 700, 13, 448, 5, 942, 2425, 563, 175, 1235, 356, 63, respectively.
	
	\subsubsection{Salinas}
	
	The Salinas dataset is another classic HSI which was also collected by the AVIRIS sensor, but over a different location in Salinas Valley, California. This dataset comprises 204 spectral bands (20 water absorption bands are removed) and $512\times 217$ pixels with a spatial resolution of 3.7 m. The Salinas dataset contains 16 land-cover classes, including `Fallow', `Stubble', and `Celery', where the numbers of examples in each class are 1979, 3696, 1946, 1364, 2648, 3929, 3549, 11241, 6173, 3248, 1038, 1897, 886, 1040, 7238, 1777, respectively.
	
	\subsection{Experimental Settings}
	
	\begin{table}[!t]
		\centering
		\caption{The Hyperparameter Settings for Different Datasets}
		\begin{tabular}{cccccc}
			\toprule
			Dataset & $\mathcal{T}$     & $\eta$ & $u$ & $s_1$ & $s_2$ \\ 
			\midrule
			Houston University  & 2000  & 0.0001 & 64 & 1 & 2\\
			Indian Pines & 2000   & 0.0001 & 128 & 1 & 4\\
			Salinas & 2000  & 0.0001 & 128 & 1 & 4 \\
			\bottomrule
		\end{tabular}%
		\label{Hyperparameters}%
	\end{table}%
	
	\begin{table*}[!t]
		\centering
		\caption{Per-Class Accuracy, OA, AA (\%), and Kappa Coefficient of Different Methods Achieved on Houston University Dataset}
		\begin{tabular}{cccccccc}
			\toprule
			Methods & S$^{2}$GCN \cite{Qin2019Spectral} & MDGCN \cite{WanShMultiscale2019} & DR-CNN \cite{Zhang2018Diverse} & CNN-PPF \cite{Li2016Hyperspectral} & MFL \cite{Li2015Multiple} & JSDF \cite{Bo2016Hyperspectral} & MGCN-AGL \\
			\midrule
			C1    & 96.30$\pm$3.07 & 93.42$\pm$4.25 & 95.62$\pm$5.41 & \textbf{98.62$\pm$0.71} & 87.23$\pm$0.47 & 97.41$\pm$1.21 & 92.25$\pm$4.77 \\
			C2    & 98.57$\pm$1.47 & 93.67$\pm$3.60 & 96.78$\pm$3.92 & 98.15$\pm$0.53 & 92.72$\pm$0.70 & \textbf{99.48$\pm$0.25} & 96.68$\pm$1.99 \\
			C3    & 98.88$\pm$0.43 & 98.12$\pm$1.09 & 96.75$\pm$1.83 & 99.01$\pm$0.33 & 99.75$\pm$0.00 & \textbf{99.88$\pm$0.22} & 97.57$\pm$1.74 \\
			C4    & 97.68$\pm$2.89 & 95.58$\pm$1.85 & 93.41$\pm$3.23 & 93.21$\pm$0.48 & 91.36$\pm$0.55 & \textbf{98.22$\pm$2.80} & 96.49$\pm$3.62 \\
			C5    & 97.66$\pm$1.12 & 99.00$\pm$1.30 & 99.15$\pm$0.78 & 99.13$\pm$0.73 & 96.83$\pm$0.36 & \textbf{100.00$\pm$0.00} & 99.89$\pm$0.25 \\
			C6    & 96.84$\pm$1.17 & 93.28$\pm$6.08 & 93.83$\pm$2.22 & 91.26$\pm$5.25 & 95.10$\pm$0.15 & \textbf{99.32$\pm$1.09} & 94.97$\pm$3.84 \\
			C7    & 83.48$\pm$5.89 & 87.68$\pm$4.41 & 80.71$\pm$6.26 & 81.54$\pm$5.84 & 86.99$\pm$0.81 & \textbf{91.93$\pm$4.91} & 91.31$\pm$4.14 \\
			C8    & 76.15$\pm$4.37 & 80.45$\pm$6.12 & 78.32$\pm$5.43 & 68.15$\pm$3.97 & 55.74$\pm$0.72 & 68.82$\pm$6.16 & \textbf{88.42$\pm$5.57} \\
			C9    & 82.17$\pm$1.78 & 89.64$\pm$2.26 & 76.90$\pm$5.62 & 77.17$\pm$1.65 & 61.38$\pm$0.56 & 69.47$\pm$8.56 & \textbf{91.24$\pm$4.94} \\
			C10   & 86.85$\pm$8.32 & 90.06$\pm$6.41 & 81.99$\pm$7.04 & 92.12$\pm$1.76 & 73.91$\pm$1.67 & 85.63$\pm$9.32 & \textbf{92.55$\pm$3.79} \\
			C11   & 88.57$\pm$5.06 & 86.73$\pm$3.22 & 84.04$\pm$4.86 & 81.05$\pm$3.31 & 85.50$\pm$0.34 & \textbf{94.51$\pm$3.82} & 87.89$\pm$6.41 \\
			C12   & 78.64$\pm$4.79 & 89.44$\pm$5.69 & 81.92$\pm$8.31 & 78.10$\pm$5.07 & 70.84$\pm$0.34 & 84.33$\pm$5.33 & \textbf{89.85$\pm$5.23} \\
			C13   & 75.62$\pm$6.93 & 92.78$\pm$4.45 & 86.54$\pm$2.58 & 72.55$\pm$4.36 & 80.06$\pm$0.63 & \textbf{98.10$\pm$1.28} & 90.09$\pm$7.02 \\
			C14   & 99.45$\pm$0.44 & 99.43$\pm$0.97 & 99.31$\pm$1.18 & 99.85$\pm$0.16 & 97.73$\pm$0.12 & \textbf{100.00$\pm$0.00} & \textbf{100.00$\pm$0.00} \\
			C15   & 98.03$\pm$1.07 & 96.27$\pm$1.72 & 99.60$\pm$0.50 & 98.60$\pm$0.34 & 98.43$\pm$0.15 & \textbf{99.86$\pm$0.36} & 95.49$\pm$2.93 \\
			\midrule
			OA    & 89.31$\pm$1.00 & 91.40$\pm$0.92 & 88.08$\pm$1.09 & 87.54$\pm$1.03 & 82.41$\pm$0.15 & 90.51$\pm$0.95 & \textbf{93.03$\pm$1.02} \\
			AA    & 90.33$\pm$1.06 & 92.37$\pm$0.89 & 89.66$\pm$0.95 & 88.57$\pm$0.77 & 84.90$\pm$0.10 & 92.46$\pm$0.75 & \textbf{93.65$\pm$0.94} \\
			Kappa & 88.44$\pm$1.08 & 90.70$\pm$1.00 & 87.10$\pm$1.18 & 86.53$\pm$1.12 & 80.97$\pm$0.16 & 89.74$\pm$1.03 & \textbf{92.46$\pm$1.10} \\
			\bottomrule
		\end{tabular}%
		\label{UHClassificationResults}%
	\end{table*}%
	
	In the experiments, our proposed MGCN-AGL algorithm is implemented via TensorFlow with Adam optimizer. For all datasets mentioned in Section \ref{dataset}, we randomly selected 30 labeled pixels per class for network training. If the corresponding class contains less than 30 pixels, 15 will be randomly chosen, leaving the remains for test. During training, 90\% of the labeled examples are utilized to learn the network parameters and the remaining 10\% are used as validation set for hyperparameter tuning. Considering that GCN usually does not require deep structure to achieve promising performance \cite{Qin2019Spectral, Gao2018Large-scale}, the number of graph convolutional layers is fixed to 2 at all levels. The selection of other hyperparameters in our MGCN-AGL, including the learning rate $\eta$, the number of iterations $\mathcal{T}$, the number of hidden units $u$, and the neighborhood scales $s_1$ and $s_2$ are shown in Table~\ref{Hyperparameters}. Meanwhile, in Section~\ref{para_sen}, the parametric sensitivity of the neighborhood scales $s_1$ and $s_2$ will be investigated in detail.
	
	In order to justify the effectiveness of our proposed MGCN-AGL, several recent state-of-the-art HSI classification methods are employed to conduct comparison. To be specific, we utilize two CNN-based methods, i.e., Diverse Region-based deep CNN (DR-CNN) \cite{Zhang2018Diverse} and CNN-Pixel-Pair Features (CNN-PPF) \cite{Li2016Hyperspectral}, together with two GCN-based methods, i.e, Spectral-Spatial Graph Convolutional Network (S$^2$GCN) \cite{Qin2019Spectral} and Multi-scale Dynamic Graph Convolutional Network (MDGCN) \cite{WanShMultiscale2019}. Additionally, the classification result of our MGCN-AGL is also compared with two traditional machine learning methods, namely, Multiple Feature Learning (MFL) \cite{Li2015Multiple} and Joint collaborative representation and SVM with Decision Fusion (JSDF) \cite{Bo2016Hyperspectral}. All the methods are conducted ten times, and the mean accuracies and standard deviations over these ten independent implementations are exhibited.

	\subsection{Classification Results}
	\label{CLassificationResult}
	
	To reveal the effectiveness of our proposed MGCN-AGL, we quantitatively and qualitatively evaluate the classification performance via comparing MGCN-AGL with the aforementioned baseline methods.

	\begin{figure*}[!t]
		\centering
		\subfigure[]{%
			\label{UHclassificationmap_gt}
			\resizebox*{1.6cm}{10cm}{\includegraphics{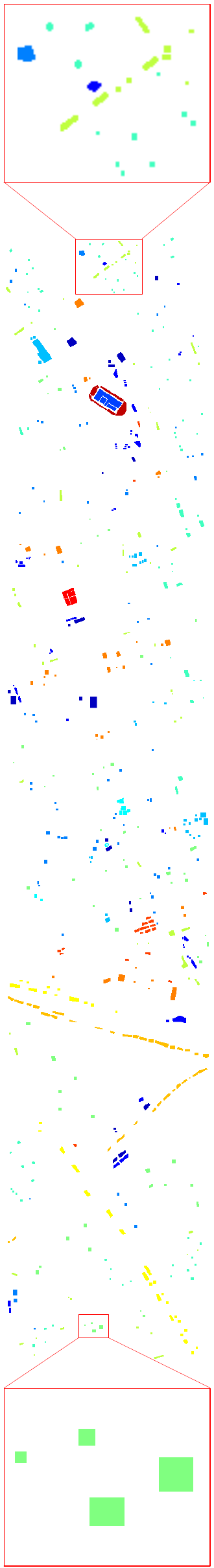}}}\hspace{5pt}	
		\subfigure[]{%
			\resizebox*{1.6cm}{10cm}{\includegraphics{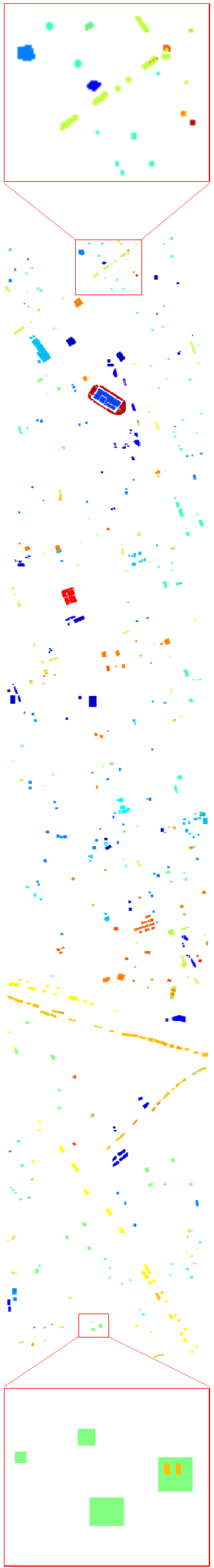}}}\hspace{5pt}
		\subfigure[]{%
			\resizebox*{1.6cm}{10cm}{\includegraphics{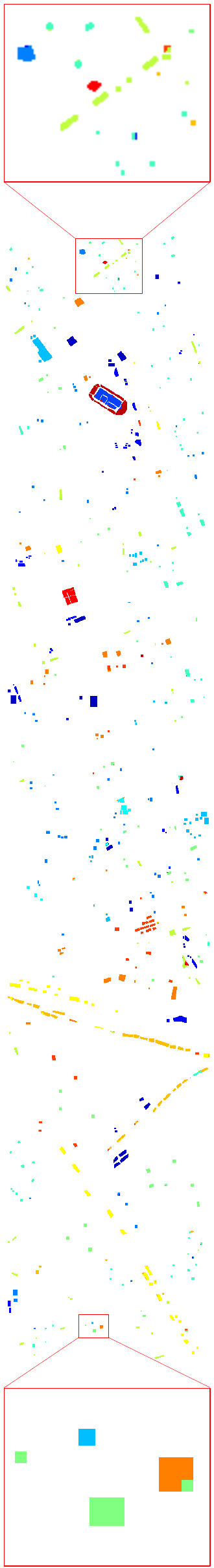}}}\hspace{5pt}	
		\subfigure[]{%
			\resizebox*{1.595cm}{10cm}{\includegraphics{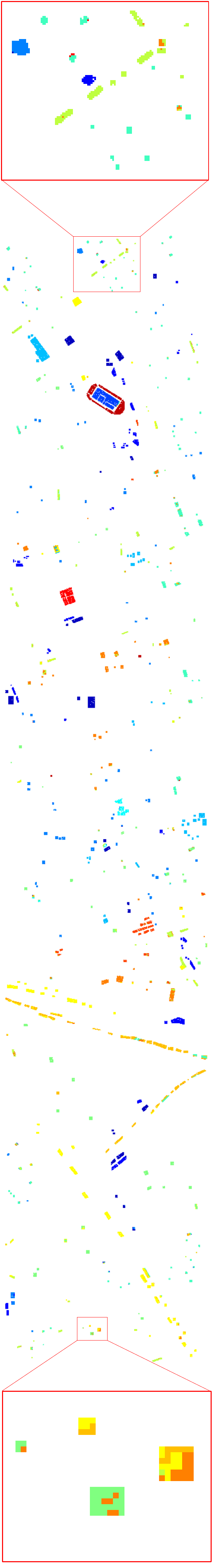}}}\hspace{5pt}		
		\subfigure[]{%
			\resizebox*{1.6cm}{10cm}{\includegraphics{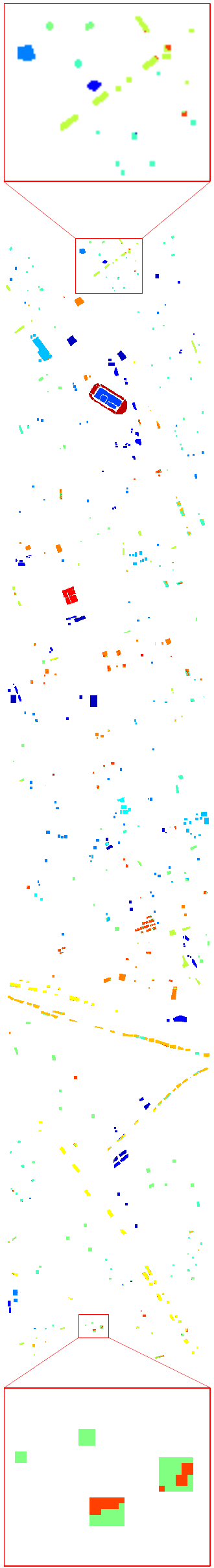}}}\hspace{5pt}
		\subfigure[]{%
			\resizebox*{1.578cm}{10cm}{\includegraphics{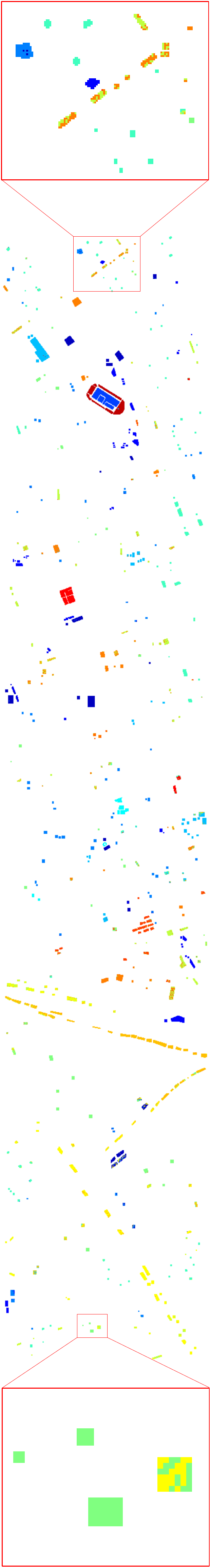}}}\hspace{5pt}		
		\subfigure[]{%
			\resizebox*{1.6cm}{10cm}{\includegraphics{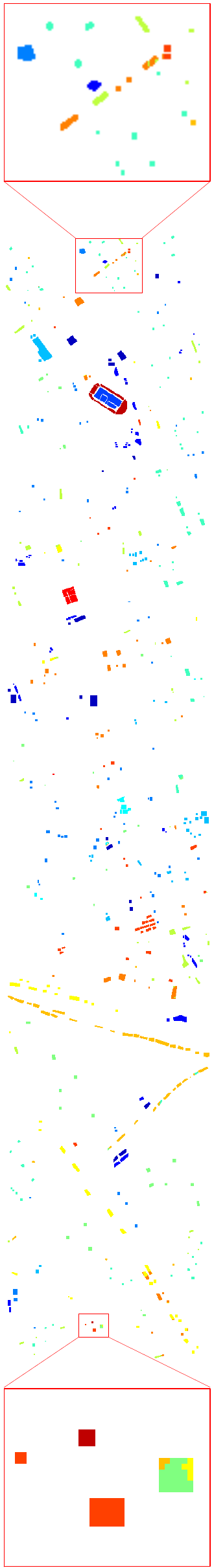}}}\hspace{5pt}	
		\subfigure[]{%
			\label{UHclassificationmap_CAD_GCN}
			\resizebox*{1.578cm}{10cm}{\includegraphics{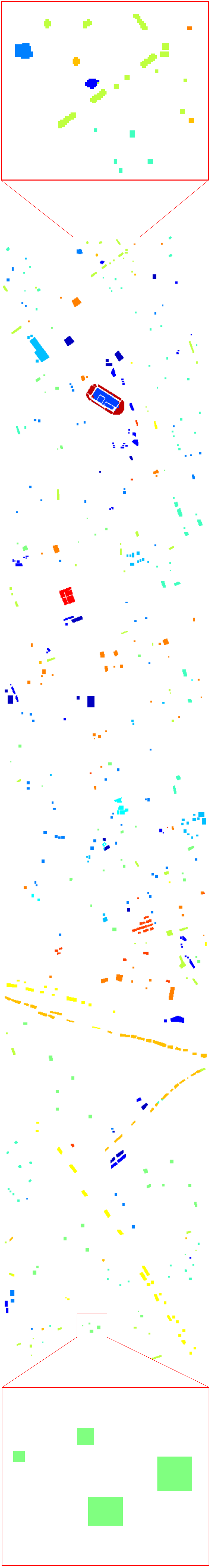}}}\hspace{0pt}
		
		\subfigure {%
			\resizebox*{!}{0.2cm}{\includegraphics{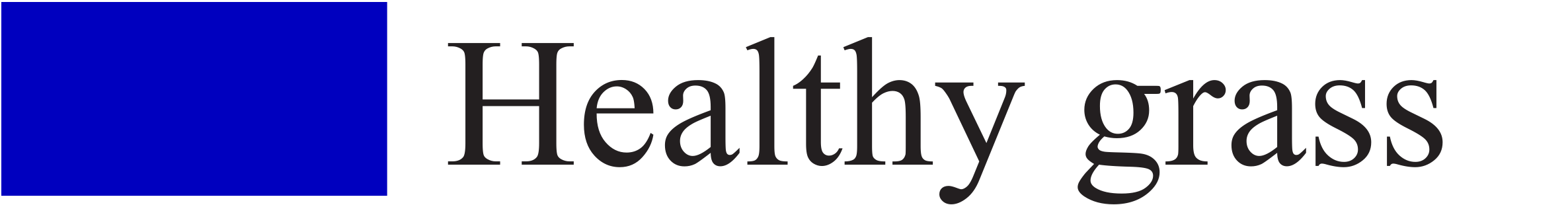}}}\hspace{2pt}
		\subfigure {%
			\resizebox*{!}{0.2cm}{\includegraphics{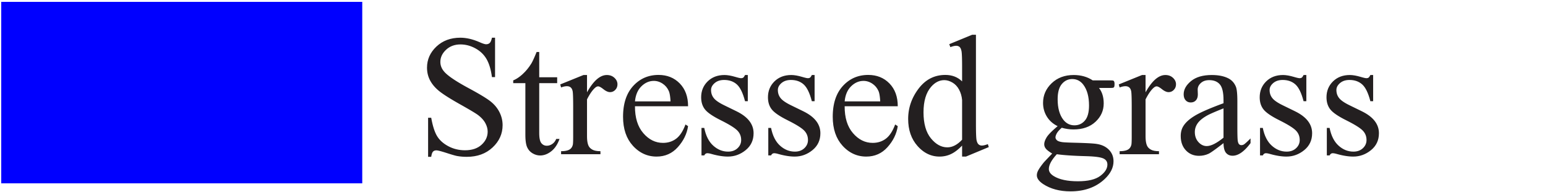}}}\hspace{2pt}
		\subfigure {%
			\resizebox*{!}{0.2cm}{\includegraphics{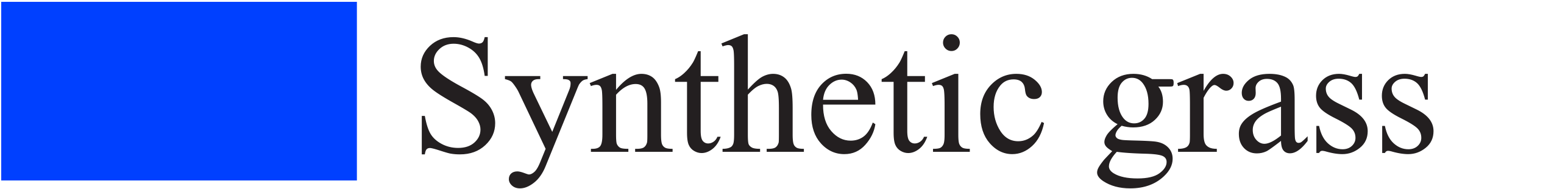}}}\hspace{3pt}
		\subfigure {%
			\resizebox*{!}{0.2cm}{\includegraphics{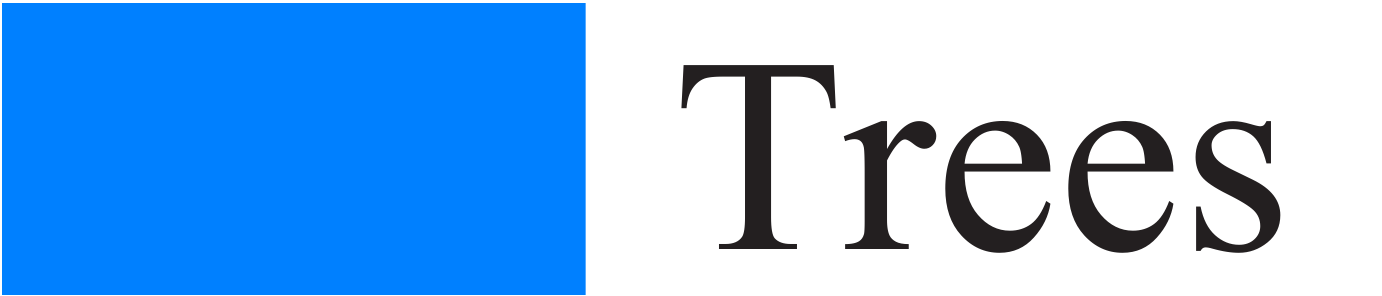}}}\hspace{5pt}
		\subfigure {%
			\resizebox*{!}{0.2cm}{\includegraphics{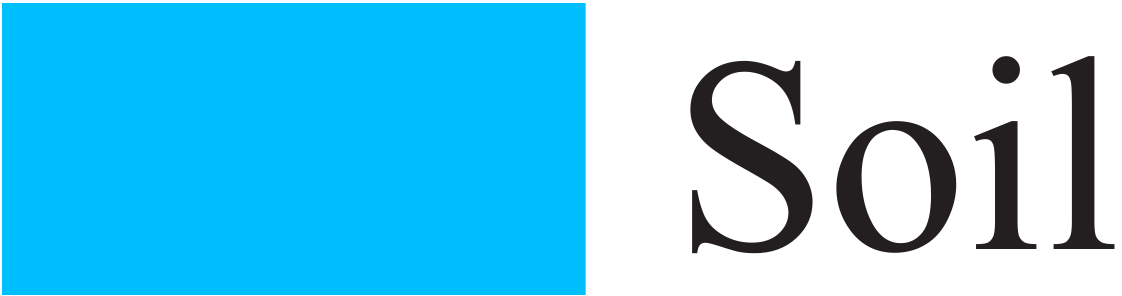}}}\hspace{5pt}
		\subfigure {%
			\resizebox*{!}{0.2cm}{\includegraphics{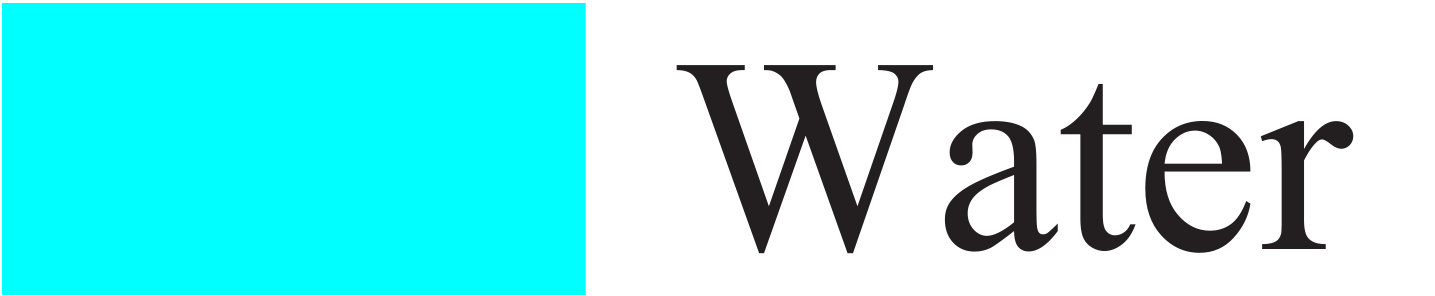}}}\hspace{4pt}	
		\subfigure {%
			\resizebox*{!}{0.2cm}{\includegraphics{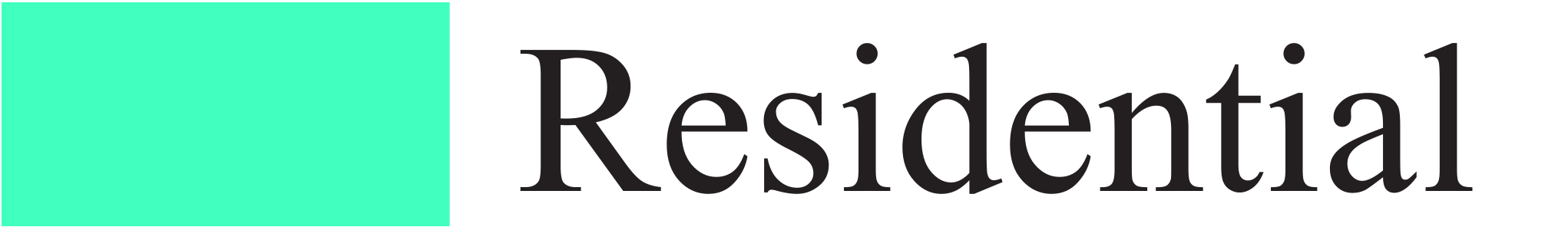}}}\hspace{5pt}
		\subfigure {%
			\resizebox*{!}{0.2cm}{\includegraphics{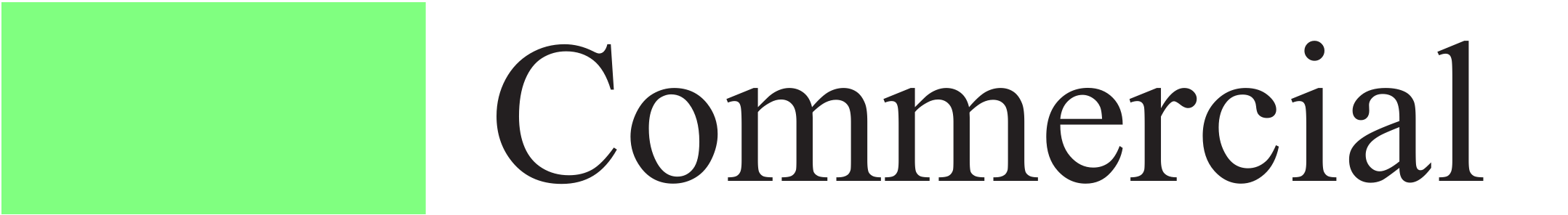}}}\hspace{5pt}
		\subfigure {%
			\resizebox*{!}{0.2cm}{\includegraphics{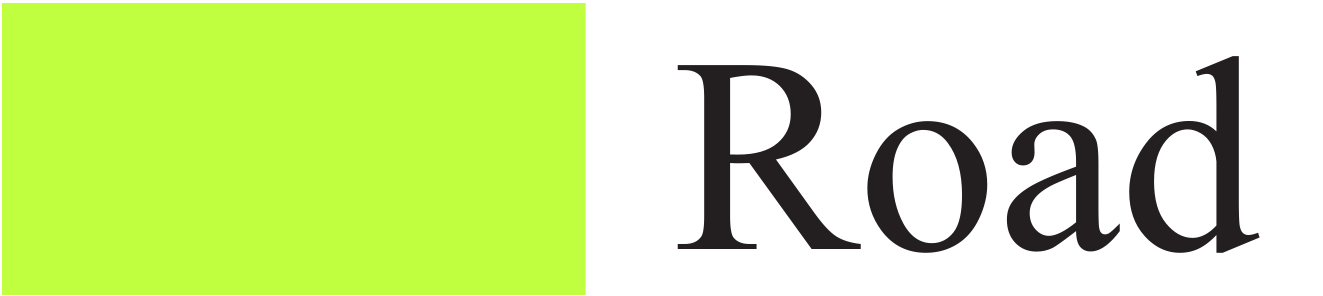}}}\hspace{5pt}	
		\subfigure {%
			\resizebox*{!}{0.2cm}{\includegraphics{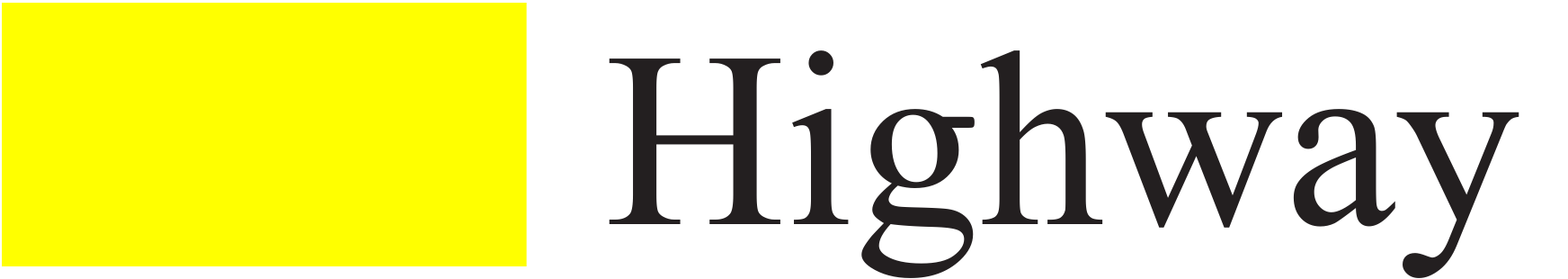}}}\hspace{5pt}
		\subfigure {%
			\resizebox*{!}{0.2cm}{\includegraphics{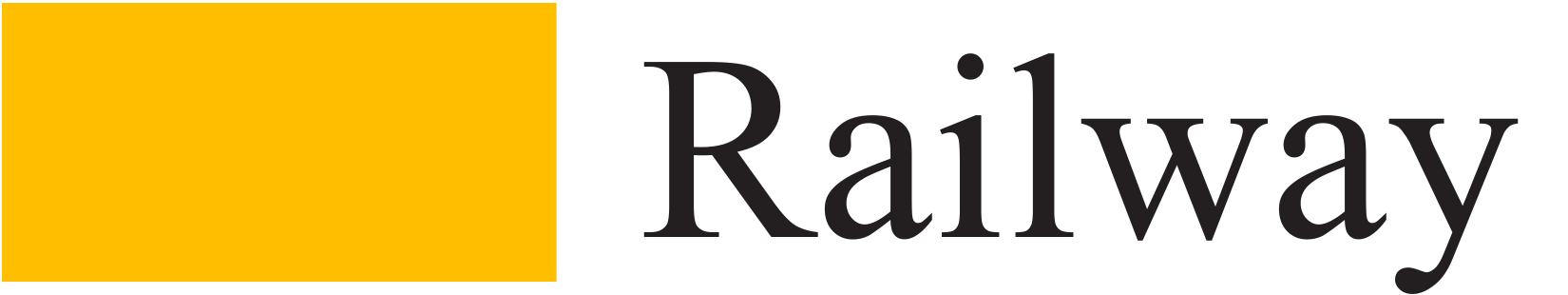}}}\hspace{5pt}
		\subfigure {%
			\resizebox*{!}{0.2cm}{\includegraphics{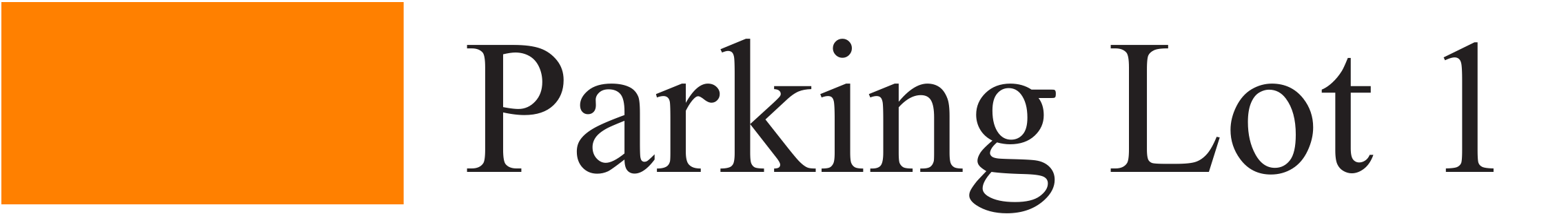}}}\hspace{2pt}		
		\subfigure {%
			\resizebox*{!}{0.2cm}{\includegraphics{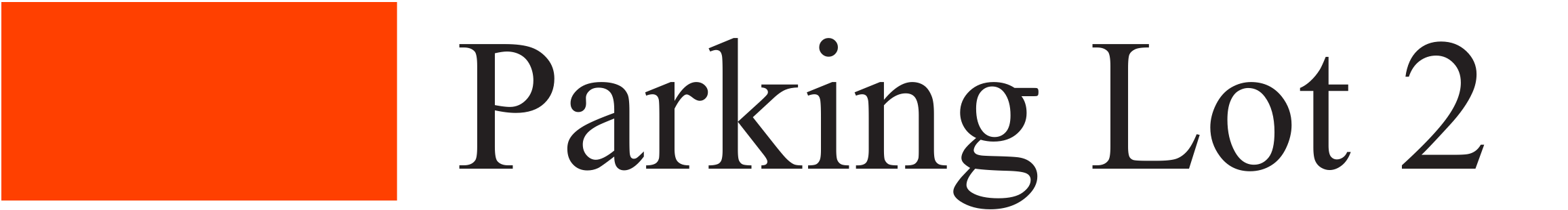}}}\hspace{4pt}	
		\subfigure {%
			\resizebox*{!}{0.2cm}{\includegraphics{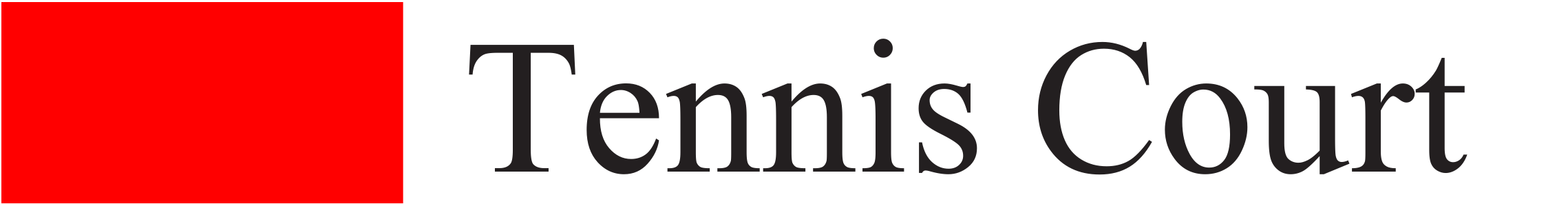}}}\hspace{2pt}
		\subfigure {%
			\resizebox*{!}{0.2cm}{\includegraphics{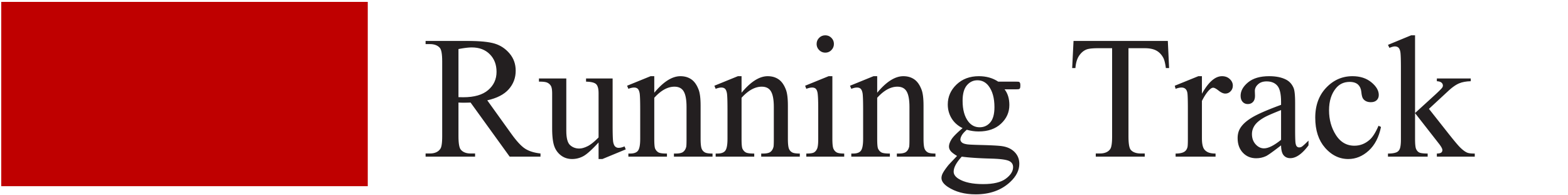}}}\hspace{2pt}	
		
		\caption{Classification maps obtained by different methods on Houston University dataset. (a) Ground truth map; (b) S$^{2}$GCN; (c) MDGCN; (d) DR-CNN; (e) CNN-PPF; (f) MFL; (g) JSDF; (h) MGCN-AGL. In (a)-(h), zoomed-in views of the regions are denoted by red boxes.} 
		\label{UHClassificationMaps}
	\end{figure*}
	
	\subsubsection{Results on the Houston University Dataset}
	
	Table~\ref{UHClassificationResults} shows the quantitative classification results achieved by different methods on the Houston University dataset, where the highest values are marked in bold in each row. For the Houston University dataset, the classes C1-C15 represent the Healthy grass, Stressed grass, Synthetic grass, Trees, Soil, Water, Residential, Commercial, Road, Highway, Railway, Parking Lot 1, Parking Lot 2, Tennis Court, and Running Track, respectively. It is noticeable that the GCN-based methods (namely, S$^{2}$GCN, MDGCN, and MGCN-AGL) outperform the CNN-based ones, which demonstrates the superiority of GCN in HSI classification. Compared with our MGCN-AGL, although JSDF (one of the traditional machine learning methods) obtains the best results in ten land-cover classes, significant decline of the accuracies can still be observed in C8 and C9. 
	
	Fig.~\ref{UHClassificationMaps} presents the a visual comparison of the classification results produced by seven different methods on the Houston University dataset, where Fig.~\ref{UHclassificationmap_gt} is the ground-truth map. As can be seen, the proposed MGCN-AGL is able to generate more precise results, and meanwhile there exist fewer errors in the zoomed-in regions, compared with other methods. Consequently, we can reasonably infer that the proposed MGCN-AGL is more effective than the compared methods.
	
	\begin{table*}[!t]
		\centering
		\caption{Per-Class Accuracy, OA, AA (\%), and Kappa Coefficient of Different Methods Achieved on Indian Pines Dataset}
		\begin{tabular}{cccccccc}
			\toprule
			Methods & S$^{2}$GCN \cite{Qin2019Spectral} & MDGCN \cite{WanShMultiscale2019} & DR-CNN \cite{Zhang2018Diverse} & CNN-PPF \cite{Li2016Hyperspectral} & MFL \cite{Li2015Multiple} & JSDF \cite{Bo2016Hyperspectral} & MGCN-AGL \\
			\midrule
			C1    & \textbf{100.00$\pm$0.00} & \textbf{100.00$\pm$0.00} & \textbf{100.00$\pm$0.00} & 95.00$\pm$2.64 & 97.64$\pm$0.88 & \textbf{100.00$\pm$0.00} & 99.55$\pm$1.44 \\
			C2    & 84.43$\pm$2.50 & 80.18$\pm$0.84 & 80.38$\pm$1.50 & 73.53$\pm$5.61 & 67.93$\pm$0.42 & \textbf{90.75$\pm$3.19} & 89.80$\pm$3.49 \\
			C3    & 82.87$\pm$5.53 & \textbf{98.26$\pm$0.00} & 82.21$\pm$3.53 & 81.34$\pm$3.76 & 71.03$\pm$0.63 & 77.84$\pm$3.81 & 96.29$\pm$2.66 \\
			C4    & 93.08$\pm$1.95 & 98.57$\pm$0.00 & 99.19$\pm$0.74 & 91.84$\pm$3.53 & 85.84$\pm$0.70 & \textbf{99.86$\pm$0.33} & 96.47$\pm$2.08 \\
			C5    & \textbf{97.13$\pm$1.34} & 95.14$\pm$0.33 & 96.47$\pm$1.10 & 93.69$\pm$0.84 & 89.36$\pm$0.48 & 87.20$\pm$2.73 & 94.25$\pm$1.88 \\
			C6    & 97.29$\pm$1.27 & 97.16$\pm$0.57 & \textbf{98.62$\pm$1.90} & 97.46$\pm$1.01 & 97.66$\pm$0.27 & 98.54$\pm$0.28 & 98.17$\pm$0.65 \\
			C7    & 92.31$\pm$0.00 & \textbf{100.00$\pm$0.00} & \textbf{100.00$\pm$0.00} & 75.38$\pm$8.73 & 95.06$\pm$0.79 & \textbf{100.00$\pm$0.00} & 86.46$\pm$16.33 \\
			C8    & 99.03$\pm$0.93 & 98.89$\pm$0.00 & 99.78$\pm$0.22 & 98.01$\pm$0.69 & 99.62$\pm$0.05 & 99.80$\pm$0.31 & \textbf{99.91$\pm$0.28} \\
			C9    & \textbf{100.00$\pm$0.00} & \textbf{100.00$\pm$0.00} & \textbf{100.00$\pm$0.00} & \textbf{100.00$\pm$0.00} & 98.00$\pm$0.94 & \textbf{100.00$\pm$0.00} & \textbf{100.00$\pm$0.00} \\
			C10   & \textbf{93.77$\pm$3.72} & 90.02$\pm$1.02 & 90.41$\pm$1.95 & 82.30$\pm$1.55 & 76.41$\pm$0.64 & 89.99$\pm$4.24 & 91.84$\pm$4.08 \\
			C11   & 84.98$\pm$2.82 & \textbf{93.35$\pm$1.47} & 74.46$\pm$0.37 & 62.64$\pm$3.32 & 73.78$\pm$0.59 & 76.75$\pm$5.12 & 91.26$\pm$2.82 \\
			C12   & 80.05$\pm$5.17 & 93.05$\pm$2.30 & 91.00$\pm$3.14 & 88.92$\pm$2.50 & 70.92$\pm$0.80 & 87.10$\pm$2.82 & \textbf{93.51$\pm$3.27} \\
			C13   & 99.43$\pm$0.00 & \textbf{100.00$\pm$0.00} & \textbf{100.00$\pm$0.00} & 98.80$\pm$0.57 & 98.80$\pm$0.08 & 99.89$\pm$0.36 & 99.72$\pm$0.88 \\
			C14   & 96.73$\pm$0.92 & \textbf{99.72$\pm$0.05} & 91.85$\pm$3.40 & 86.49$\pm$2.23 & 90.12$\pm$0.53 & 97.21$\pm$2.78 & 99.51$\pm$0.26 \\
			C15   & 86.80$\pm$3.42 & \textbf{99.72$\pm$0.00} & 99.44$\pm$0.28 & 86.71$\pm$4.36 & 96.05$\pm$0.35 & 99.58$\pm$0.68 & 98.00$\pm$3.62 \\
			C16   & \textbf{100.00$\pm$0.00} & 95.71$\pm$0.00 & \textbf{100.00$\pm$0.00} & 92.70$\pm$3.45 & 97.54$\pm$0.23 & \textbf{100.00$\pm$0.00} & 94.59$\pm$3.58 \\
			\midrule
			OA    & 89.49$\pm$1.08 & 93.47$\pm$0.38 & 86.65$\pm$0.59 & 80.09$\pm$1.56 & 80.22$\pm$0.20 & 88.34$\pm$1.39 & \textbf{94.27$\pm$0.92} \\
			AA    & 92.99$\pm$1.04 & \textbf{96.24$\pm$0.21} & 93.99$\pm$0.25 & 87.80$\pm$1.53 & 87.85$\pm$0.19 & 94.03$\pm$0.55 & 95.58$\pm$1.18 \\
			Kappa & 88.00$\pm$1.23 & 92.55$\pm$0.43 & 84.88$\pm$0.67 & 77.52$\pm$1.74 & 77.59$\pm$0.22 & 86.80$\pm$1.55 & \textbf{93.46$\pm$1.04} \\
			\bottomrule
		\end{tabular}%
		\label{IPClassificationResults}%
	\end{table*}%

	\begin{figure*}[!t]
		\centering
		\subfigure[]{%
			\label{IPClassificationMaps_gt}
			\resizebox*{3cm}{!}{\includegraphics{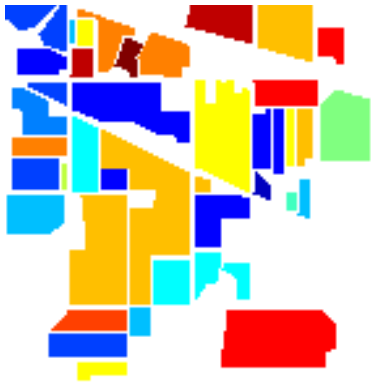}}}\hspace{10pt}
		\subfigure[]{%
			\resizebox*{3cm}{!}{\includegraphics{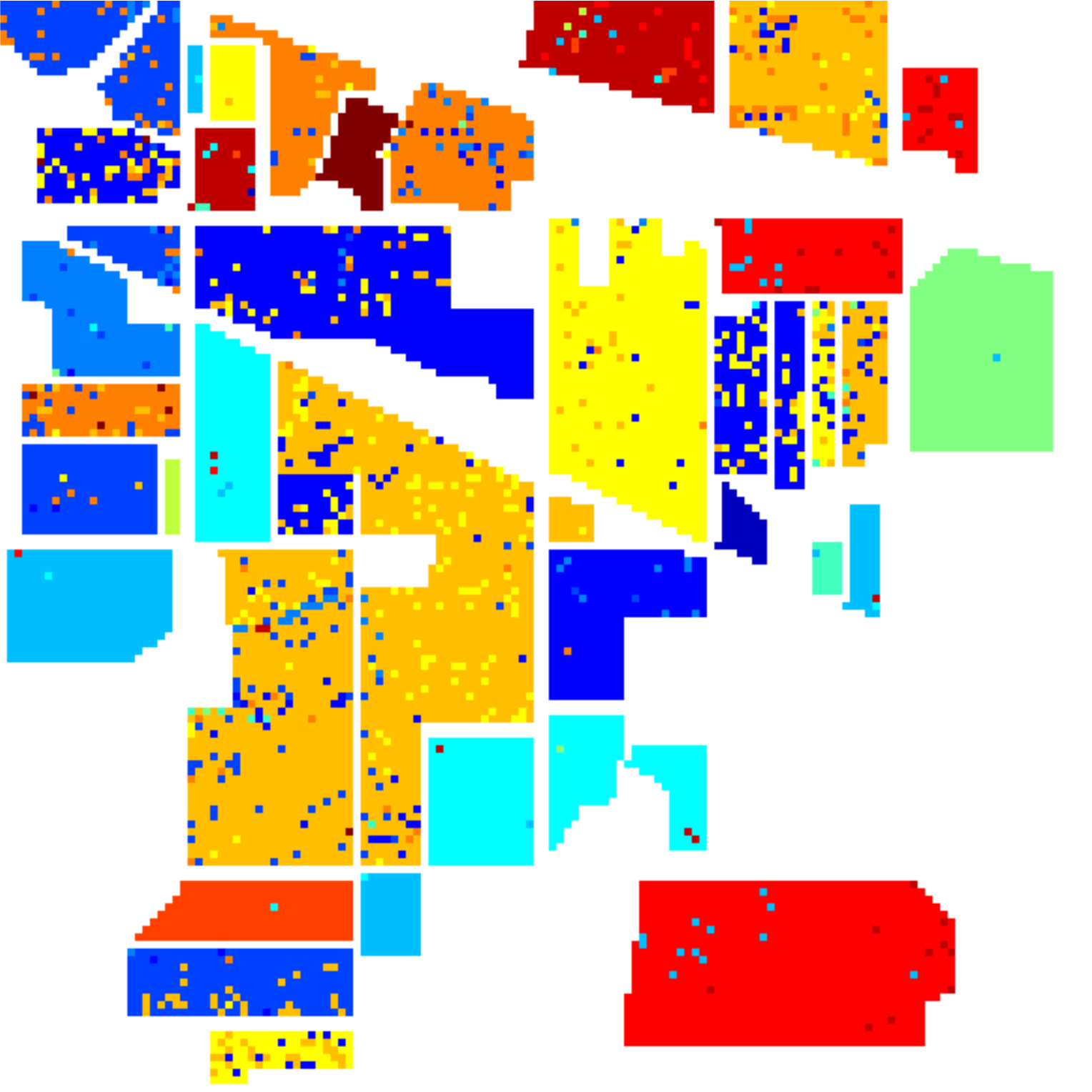}}}\hspace{10pt}	
		\subfigure[]{%
			\resizebox*{3cm}{!}{\includegraphics{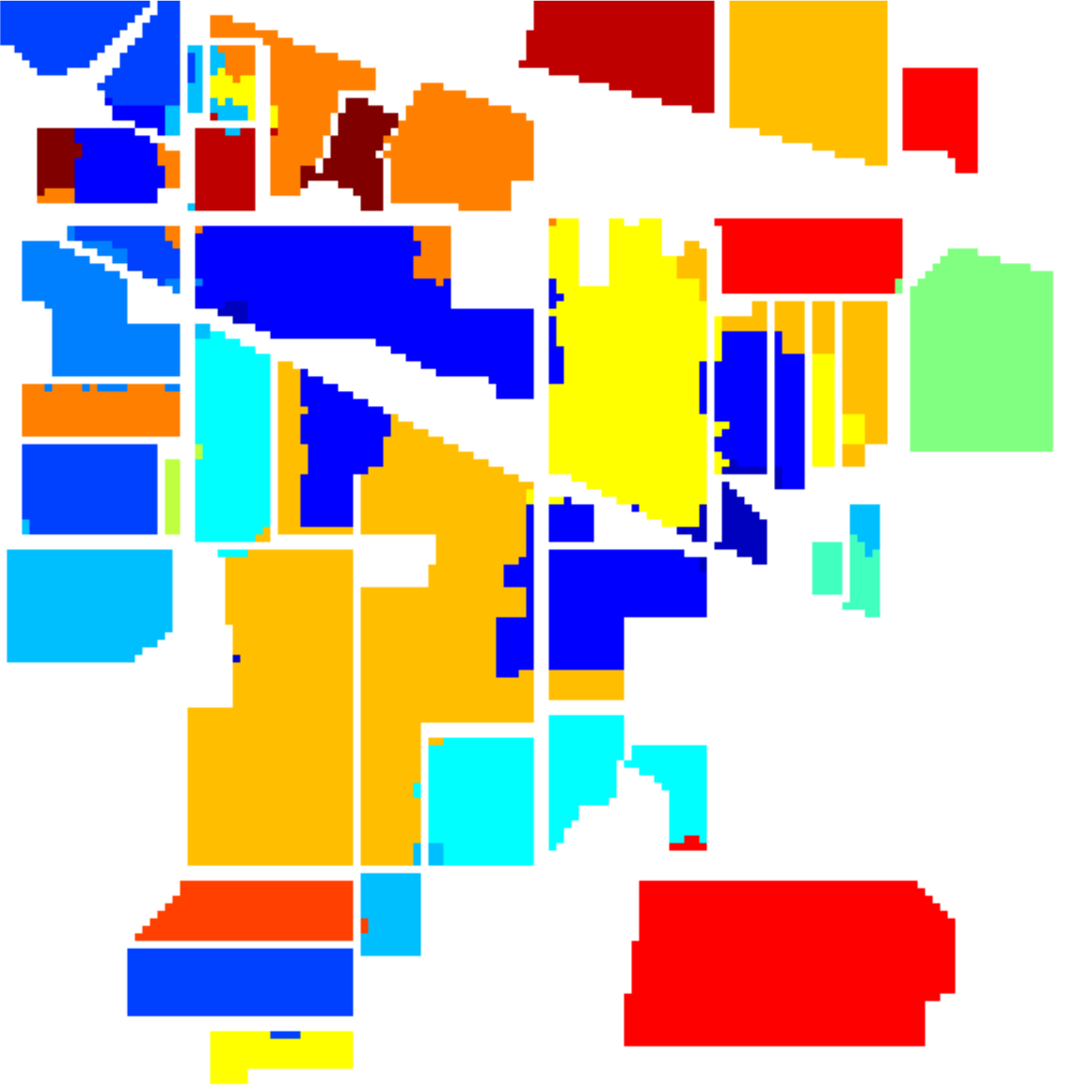}}}\hspace{10pt}	
		\subfigure[]{%
			\resizebox*{3cm}{!}{\includegraphics{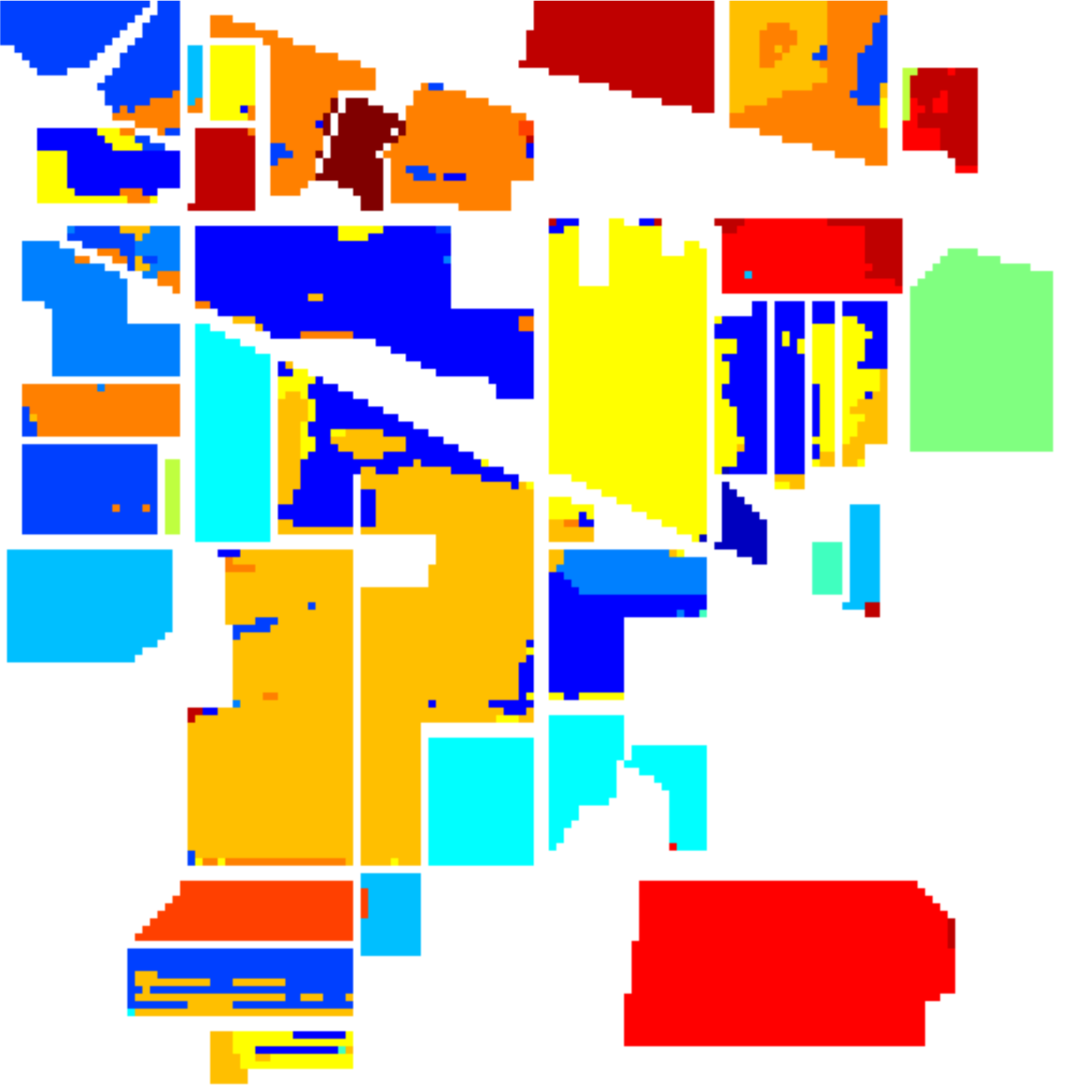}}}\hspace{0pt}
		
		\subfigure[]{%
			\resizebox*{3cm}{!}{\includegraphics{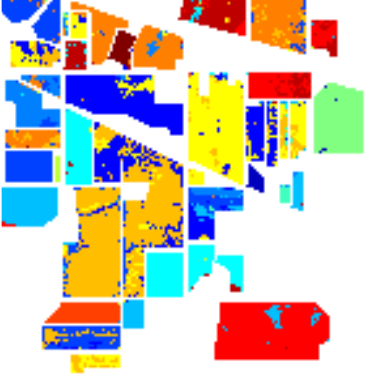}}}\hspace{10pt}	
		\subfigure[]{%
			\resizebox*{3cm}{!}{\includegraphics{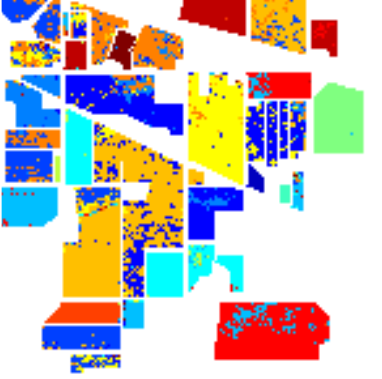}}}\hspace{10pt}			
		\subfigure[]{%
			\resizebox*{3cm}{!}{\includegraphics{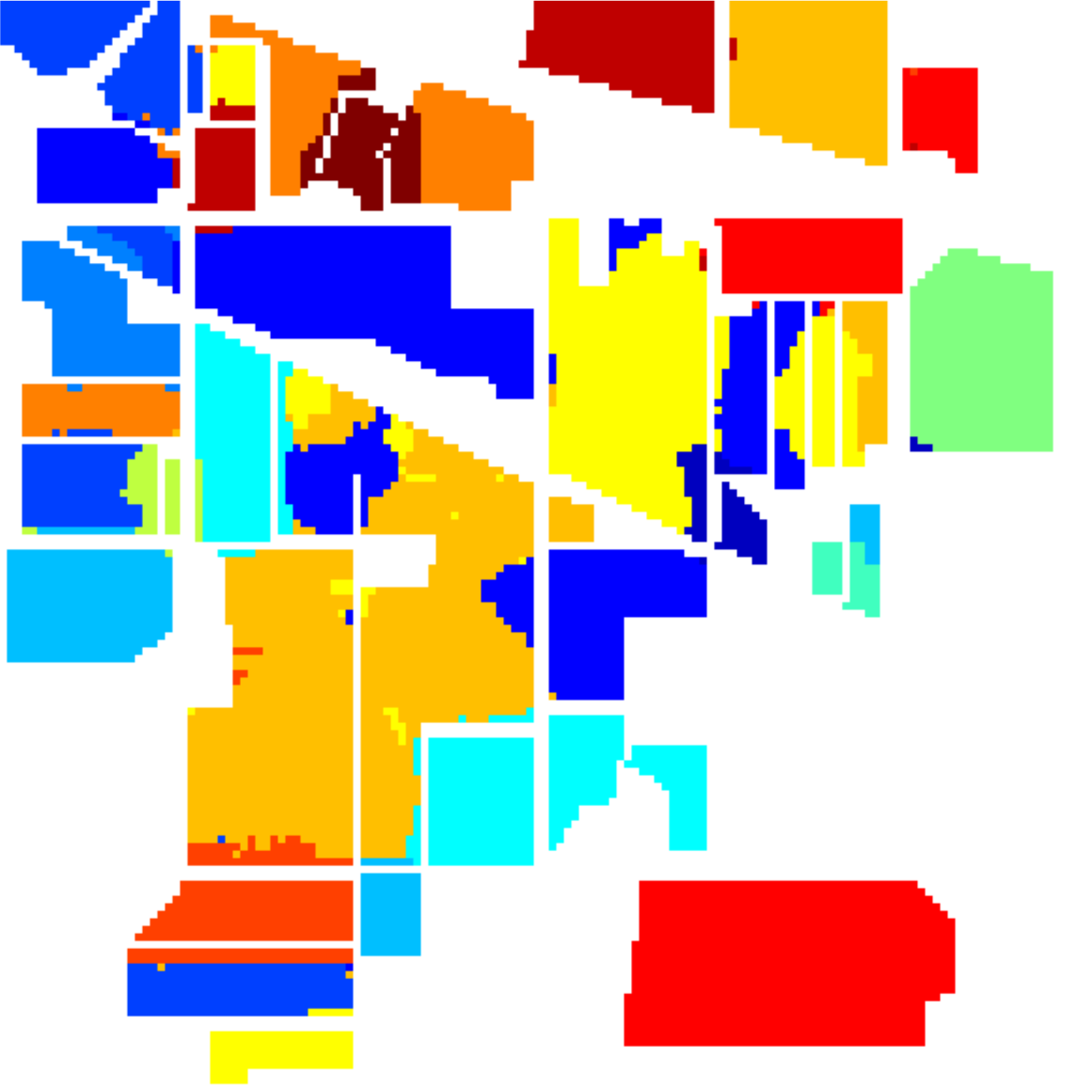}}}\hspace{10pt}	
		\subfigure[]{%
			\resizebox*{3cm}{!}{\includegraphics{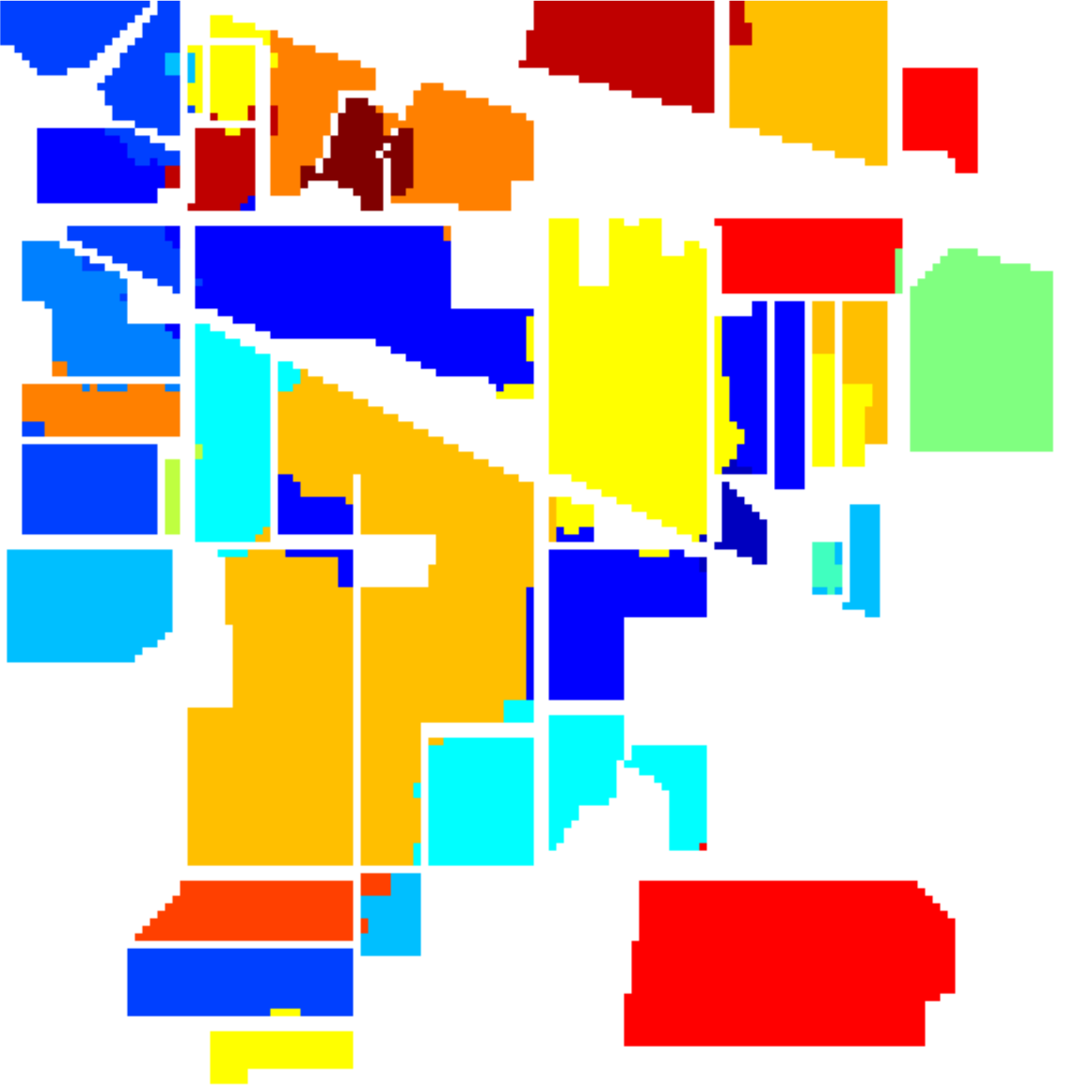}}}\hspace{0pt}	
		
		\subfigure {%
			\resizebox*{!}{0.2cm}{\includegraphics{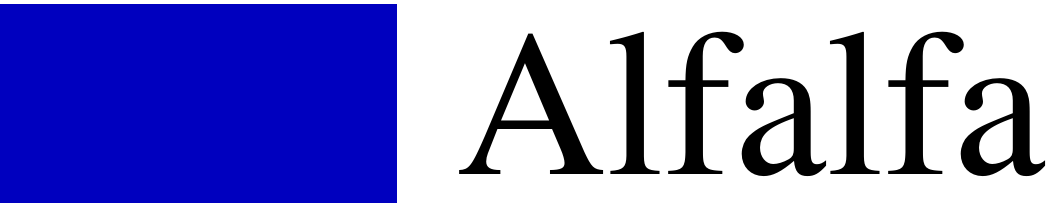}}}\hspace{9pt}
		\subfigure {%
			\resizebox*{!}{0.2cm}{\includegraphics{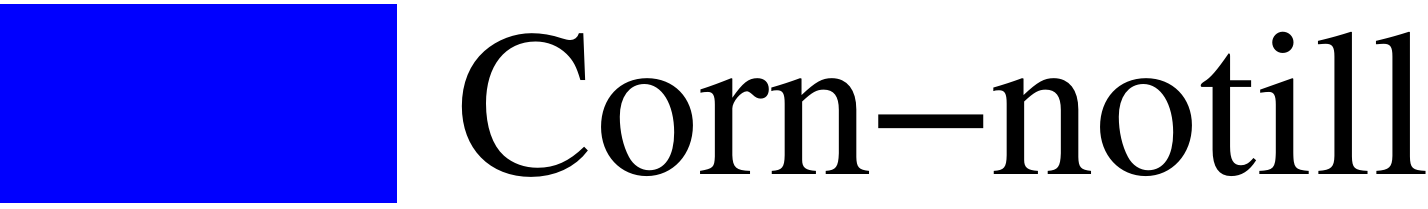}}}\hspace{9pt}
		\subfigure {%
			\resizebox*{!}{0.2cm}{\includegraphics{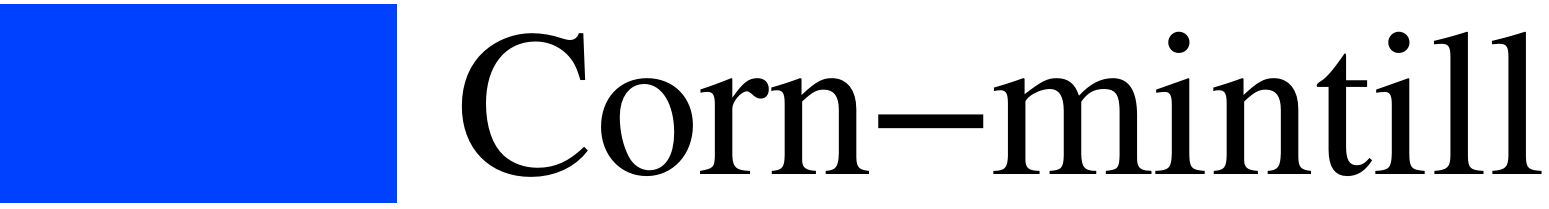}}}\hspace{9pt}
		\subfigure {%
			\resizebox*{!}{0.2cm}{\includegraphics{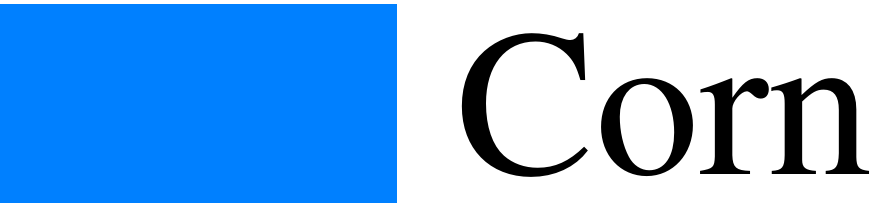}}}\hspace{9pt}
		\subfigure {%
			\resizebox*{!}{0.2cm}{\includegraphics{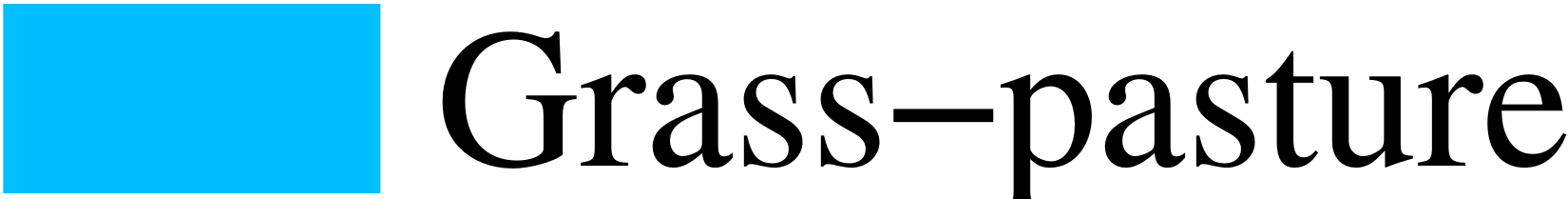}}}\hspace{9pt}
		\subfigure {%
			\resizebox*{!}{0.2cm}{\includegraphics{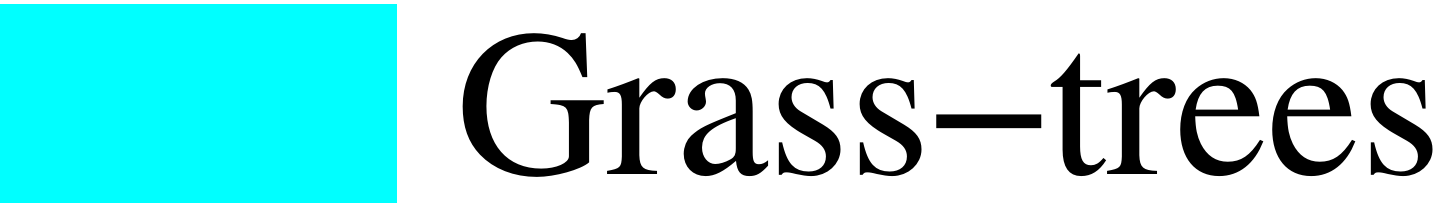}}}\hspace{9pt}
		\subfigure {%
			\resizebox*{!}{0.2cm}{\includegraphics{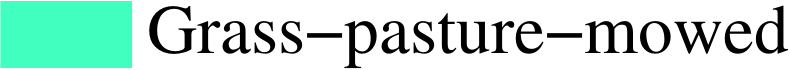}}}\hspace{9pt}
		\subfigure {%
			\resizebox*{!}{0.2cm}{\includegraphics{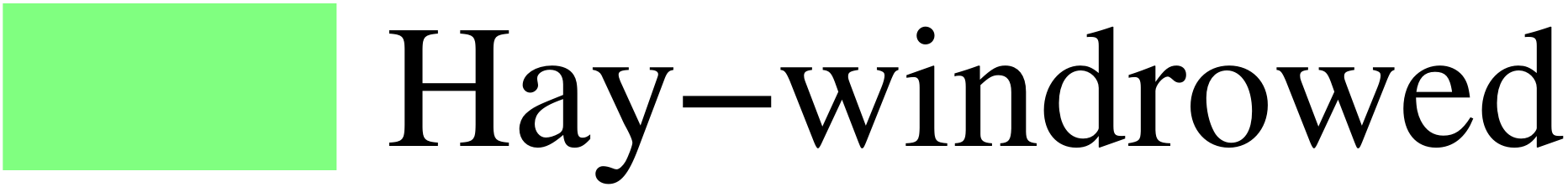}}}\hspace{9pt}
		
		\subfigure {%
			\resizebox*{!}{0.2cm}{\includegraphics{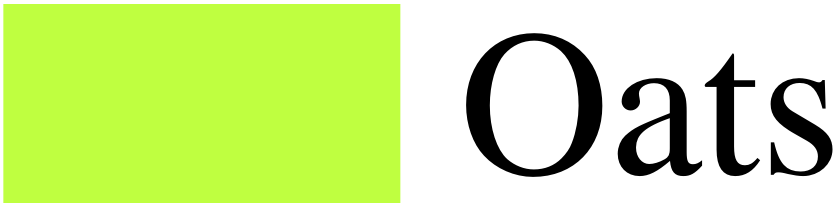}}}\hspace{9pt}
		\subfigure {%
			\resizebox*{!}{0.2cm}{\includegraphics{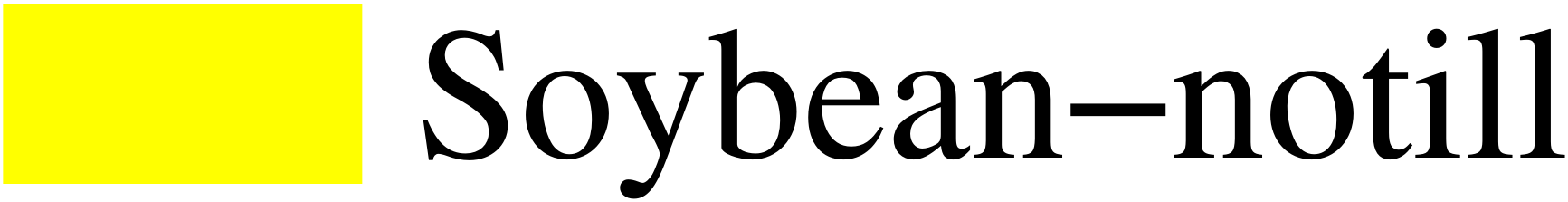}}}\hspace{9pt}
		\subfigure {%
			\resizebox*{!}{0.2cm}{\includegraphics{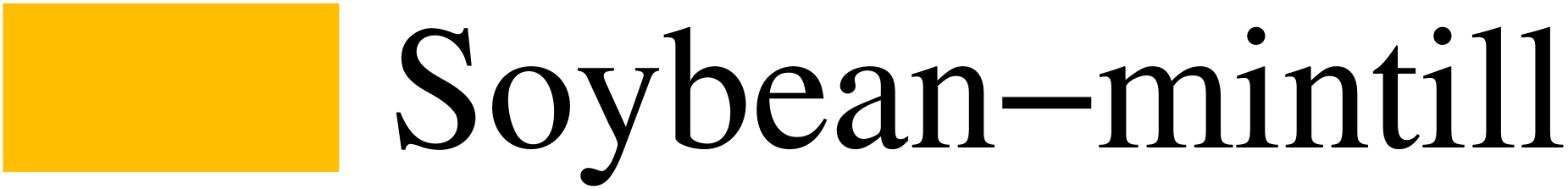}}}\hspace{9pt}
		\subfigure {%
			\resizebox*{!}{0.2cm}{\includegraphics{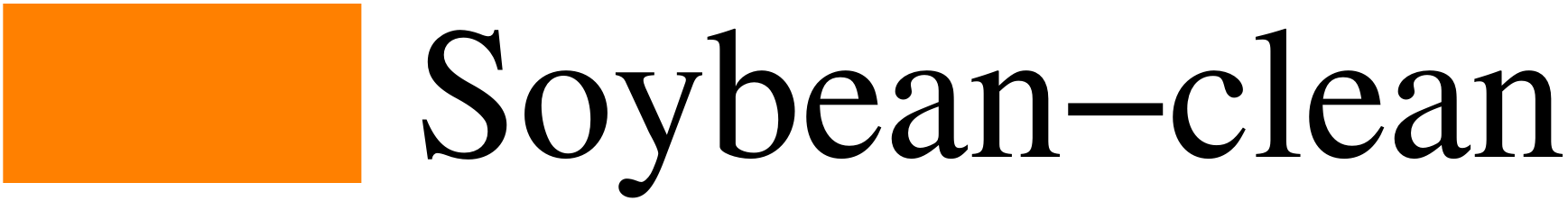}}}\hspace{9pt}
		\subfigure {%
			\resizebox*{!}{0.19cm}{\includegraphics{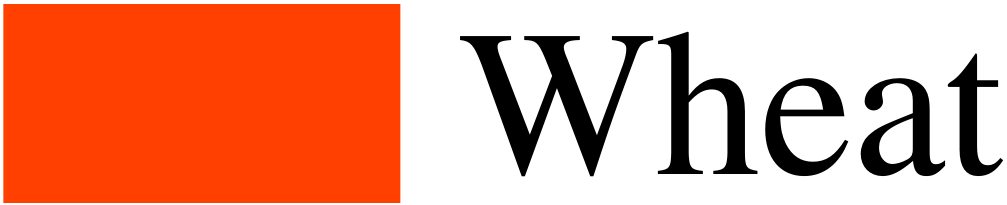}}}\hspace{9pt}
		\subfigure {%
			\resizebox*{!}{0.19cm}{\includegraphics{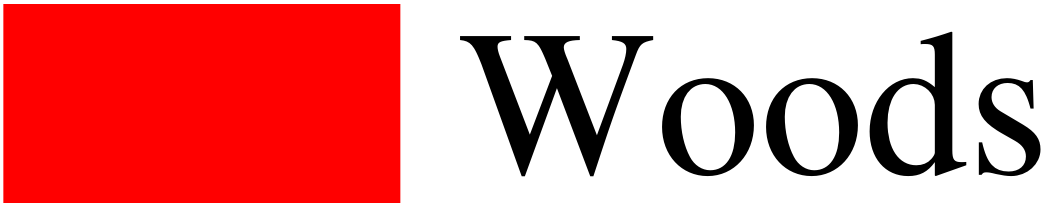}}}\hspace{9pt}
		\subfigure {%
			\resizebox*{!}{0.23cm}{\includegraphics{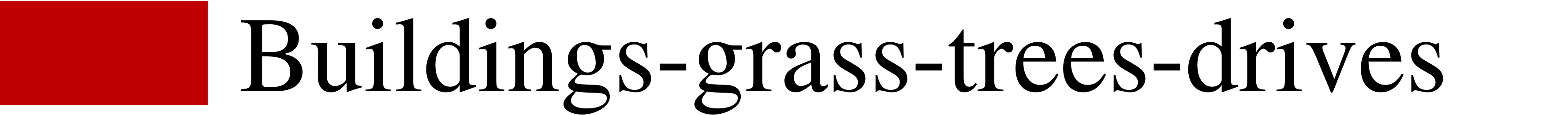}}}\hspace{4pt}
		\subfigure {%
			\resizebox*{!}{0.23cm}{\includegraphics{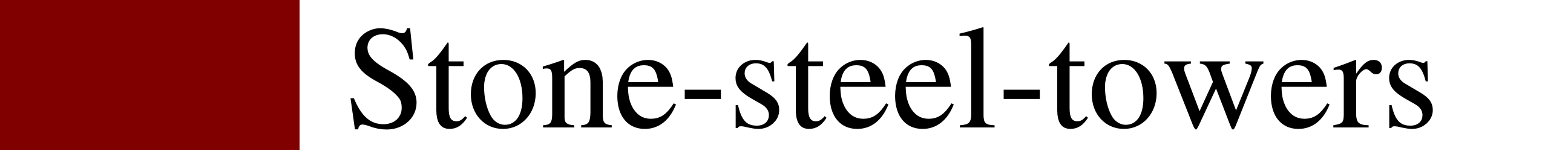}}}\hspace{0pt}
		\caption{Classification maps obtained by different methods on Indian Pines dataset. (a) Ground truth map; (b) S$^{2}$GCN; (c) MDGCN; (d) DR-CNN; (e) CNN-PPF; (f) MFL; (g) JSDF; (h) MGCN-AGL.} 
		\label{IPClassificationMaps}
	\end{figure*}	
	
	\subsubsection{Results on the Indian Pines Dataset}
	
	Table~\ref{IPClassificationResults} reports the quantitative results of different methods on the Indian Pines dataset. For this dataset, C1-C16 stand for the Alfalfa, Corn-notill, Corn-mintill, Corn, Grass-pasture, Grass-trees, Grass-pasture-mowed, Hay-windrowed, Oats, Soybean-notill, Soybean-mintill, Soybean-clean, Wheat, Woods, Buildings-grass-trees-drives, and Stone-steel-towers, respectively. We can observe that the proposed MGCN-AGL acquires very competitive results when compared with the baseline methods, which again validates the power of our proposed multi-level graph convolution with automatic graph learning. It is also noteworthy that MDGCN and our MGCN-AGL outperform the other methods by a substantial margin in terms of OA, and the standard deviations are relatively small as well. It can be inferred that the multi-level spatial context is critical in obtaining promising performance for HSI classification.
	
	Fig.~\ref{IPClassificationMaps} visualizes the results generated via using different methods, where the classification map obtained by our proposed MGCN-AGL is noticeably closer to the ground-truth map (see Fig.~\ref{IPClassificationMaps_gt}) than those of other methods. In addition, we can observe that the GCN-based methods (namely, S$^{2}$GCN, MDGCN, and MGCN-AGL) produce fewer errors around class boundaries than the other ones, which confirms the good discriminability of the GCN in boundary regions
	
	\begin{table*}[!t]
		\centering
		\caption{Per-Class Accuracy, OA, AA (\%), and Kappa Coefficient of Different Methods Achieved on Salinas Dataset}
		\begin{tabular}{cccccccc}
			\toprule
			Methods & S$^{2}$GCN \cite{Qin2019Spectral} & MDGCN \cite{WanShMultiscale2019} & DR-CNN \cite{Zhang2018Diverse} & CNN-PPF \cite{Li2016Hyperspectral} & MFL \cite{Li2015Multiple} & JSDF \cite{Bo2016Hyperspectral} & MGCN-AGL \\
			\midrule
			C1    & 99.01$\pm$0.44 & 99.98$\pm$0.03 & 99.40$\pm$1.54 & 99.77$\pm$0.21 & 98.41$\pm$0.09 & \textbf{100.00$\pm$0.00} & \textbf{100.00$\pm$0.00} \\
			C2    & 99.18$\pm$0.59 & 99.90$\pm$0.28 & 99.46$\pm$0.16 & 98.69$\pm$0.89 & 99.04$\pm$0.06 & \textbf{100.00$\pm$0.00} & \textbf{100.00$\pm$0.00} \\
			C3    & 97.15$\pm$2.76 & 99.80$\pm$0.21 & 98.58$\pm$1.69 & 99.50$\pm$0.49 & 99.74$\pm$0.04 & \textbf{100.00$\pm$0.00} & 99.97$\pm$0.07 \\
			C4    & 99.11$\pm$0.55 & 97.49$\pm$2.16 & 99.70$\pm$0.45 & 99.81$\pm$0.04 & 98.43$\pm$0.14 & \textbf{99.93$\pm$0.09} & 98.40$\pm$0.44 \\
			C5    & 97.55$\pm$2.35 & 97.96$\pm$0.77 & 98.90$\pm$0.74 & 96.64$\pm$1.26 & 98.53$\pm$0.02 & \textbf{99.77$\pm$0.31} & 95.21$\pm$1.66 \\
			C6    & 99.32$\pm$0.35 & 99.10$\pm$1.67 & 99.57$\pm$0.78 & 99.32$\pm$0.86 & 98.97$\pm$0.11 & \textbf{100.00$\pm$0.00} & 99.67$\pm$0.19 \\
			C7    & 99.06$\pm$0.27 & 98.18$\pm$1.49 & 99.50$\pm$0.66 & 99.59$\pm$0.13 & 99.14$\pm$0.03 & \textbf{99.99$\pm$0.01} & 98.93$\pm$2.54 \\
			C8    & 70.68$\pm$5.20 & 92.78$\pm$4.61 & 75.59$\pm$8.19 & 74.77$\pm$4.01 & 69.74$\pm$0.86 & 87.79$\pm$4.89 & \textbf{97.38$\pm$3.26} \\
			C9    & 98.32$\pm$1.79 & \textbf{100.00$\pm$0.00} & 99.75$\pm$0.41 & 98.99$\pm$0.18 & 98.95$\pm$0.04 & 99.67$\pm$0.33 & \textbf{100.00$\pm$0.00} \\
			C10   & 90.97$\pm$2.59 & 98.31$\pm$1.29 & 94.29$\pm$2.24 & 89.32$\pm$3.04 & 90.66$\pm$0.29 & 96.53$\pm$2.55 & \textbf{98.54$\pm$0.70} \\
			C11   & 98.00$\pm$1.65 & 99.39$\pm$0.55 & 97.57$\pm$2.19 & 97.65$\pm$1.49 & 93.85$\pm$0.28 & \textbf{99.76$\pm$0.21} & 99.65$\pm$0.21 \\
			C12   & 99.56$\pm$0.59 & 99.01$\pm$0.78 & 99.99$\pm$0.05 & 99.82$\pm$0.30 & 97.85$\pm$0.31 & \textbf{100.00$\pm$0.00} & 98.16$\pm$1.19 \\
			C13   & 97.83$\pm$0.72 & 97.59$\pm$1.32 & 99.95$\pm$0.09 & 97.70$\pm$0.50 & 99.12$\pm$0.10 & \textbf{100.00$\pm$0.00} & 96.87$\pm$1.08 \\
			C14   & 95.75$\pm$1.65 & 97.92$\pm$1.72 & 98.57$\pm$1.13 & 94.14$\pm$1.22 & 94.52$\pm$0.32 & \textbf{98.71$\pm$0.72} & 98.10$\pm$1.42 \\
			C15   & 70.36$\pm$3.62 & 95.71$\pm$4.57 & 72.18$\pm$9.28 & 79.12$\pm$1.99 & 71.09$\pm$0.83 & 81.86$\pm$5.26 & \textbf{96.79$\pm$2.46} \\
			C16   & 96.90$\pm$1.97 & 98.18$\pm$2.92 & 98.45$\pm$0.57 & 98.65$\pm$0.31 & 99.37$\pm$0.05 & 98.99$\pm$0.63 & \textbf{100.00$\pm$0.00} \\
			\midrule
			OA    & 88.39$\pm$1.01 & 97.25$\pm$0.87 & 90.35$\pm$1.14 & 90.52$\pm$0.77 & 88.36$\pm$0.22 & 94.67$\pm$0.77 & \textbf{98.39$\pm$0.63} \\
			AA    & 94.30$\pm$0.47 & 98.21$\pm$0.30 & 95.72$\pm$0.39 & 95.22$\pm$0.34 & 94.21$\pm$0.08 & 97.69$\pm$0.34 & \textbf{98.60$\pm$0.24} \\
			Kappa & 87.10$\pm$1.12 & 96.94$\pm$0.96 & 89.26$\pm$1.26 & 89.46$\pm$0.85 & 87.06$\pm$0.24 & 94.06$\pm$0.85 & \textbf{98.21$\pm$0.70} \\
			\bottomrule
		\end{tabular}%
		\label{SAClassificationResults}%
	\end{table*}%
	
	\begin{figure*}[!t]
		\centering
		\subfigure[]{%
			\label{SA_classified_gt}
			\resizebox*{2.2cm}{4.4cm}{\includegraphics{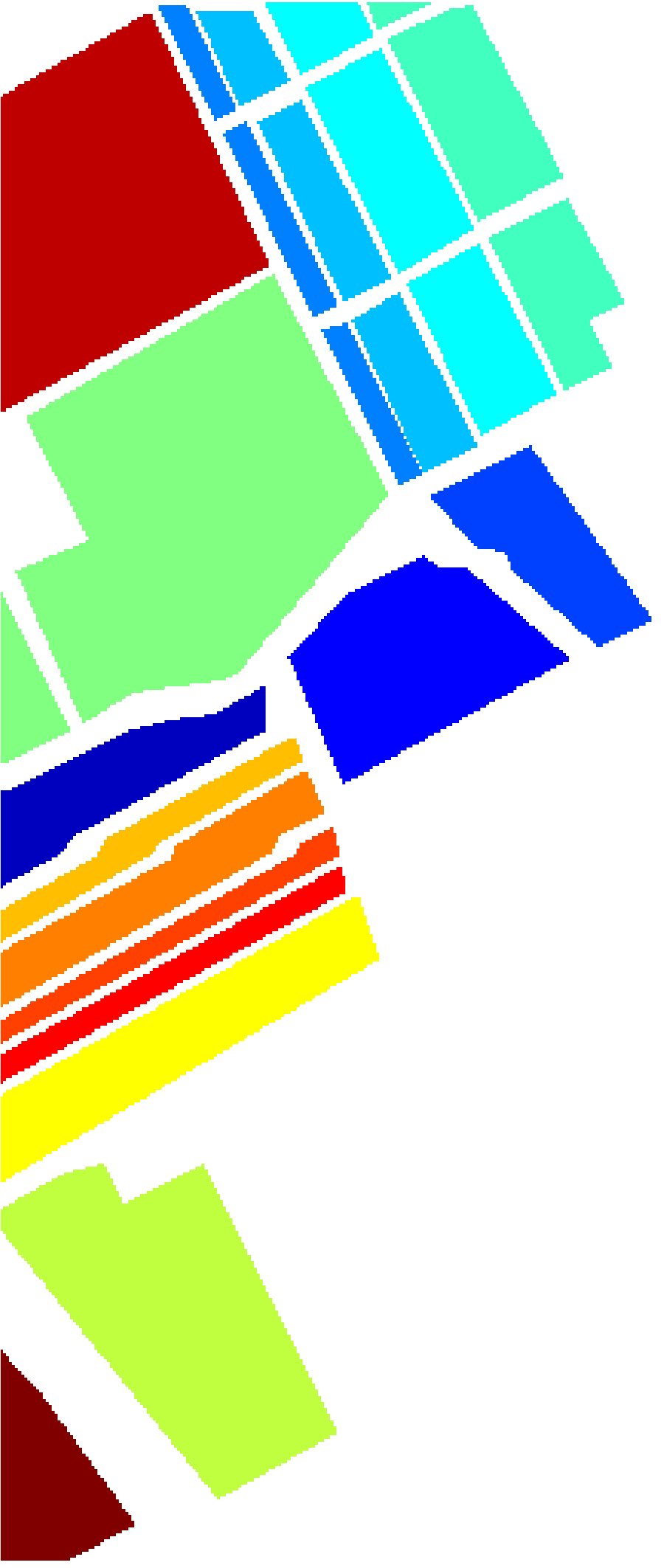}}}\hspace{40pt}
		\subfigure[]{%
			\resizebox*{2.2cm}{4.4cm}{\includegraphics{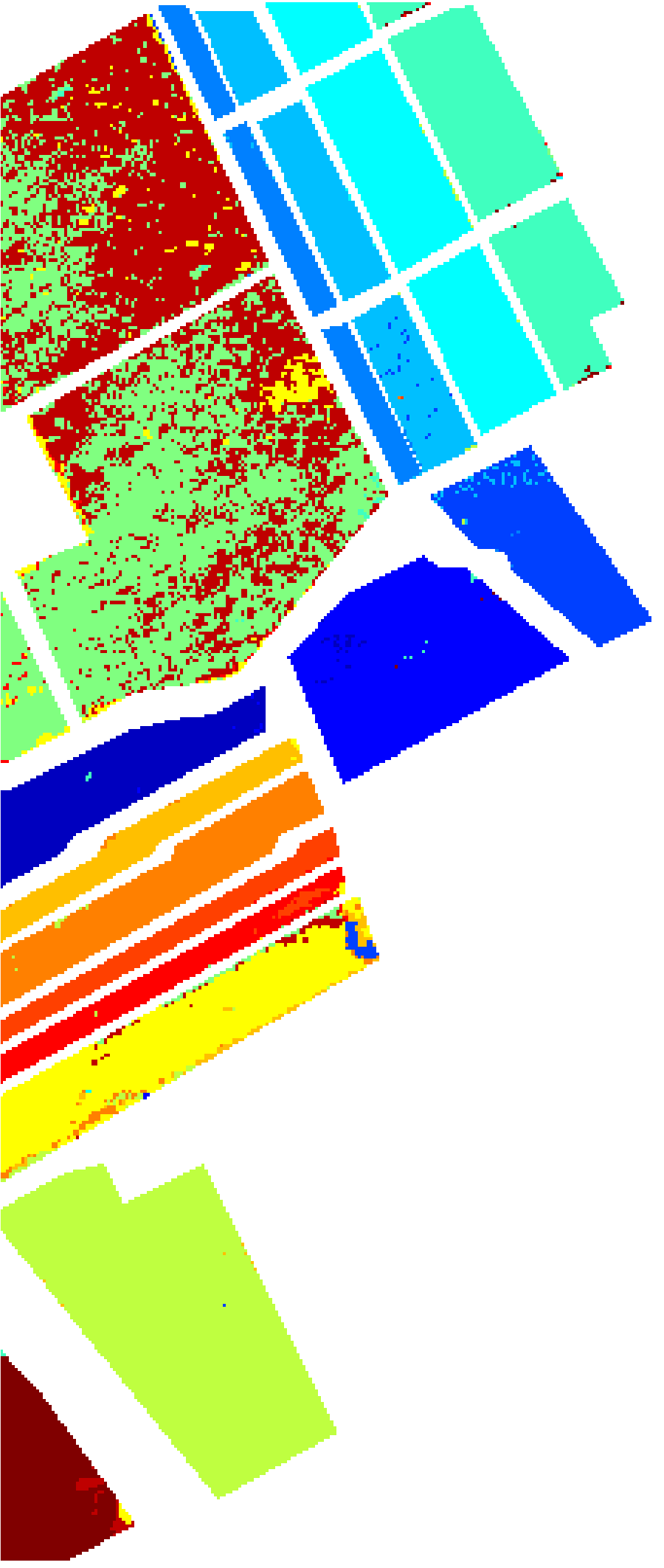}}}\hspace{40pt}	
		\subfigure[]{%
			\resizebox*{2.2cm}{4.4cm}{\includegraphics{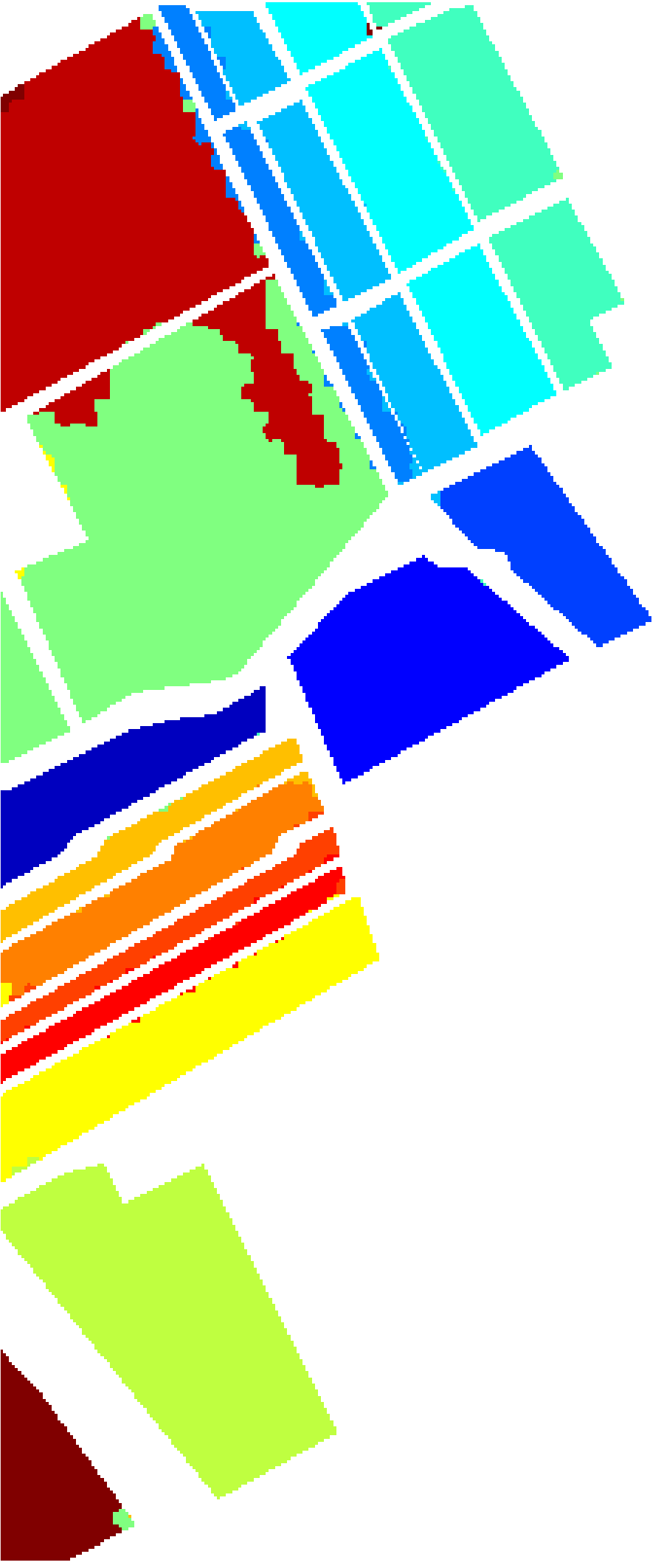}}}\hspace{40pt}	
		\subfigure[]{%
			\resizebox*{2.2cm}{4.4cm}{\includegraphics{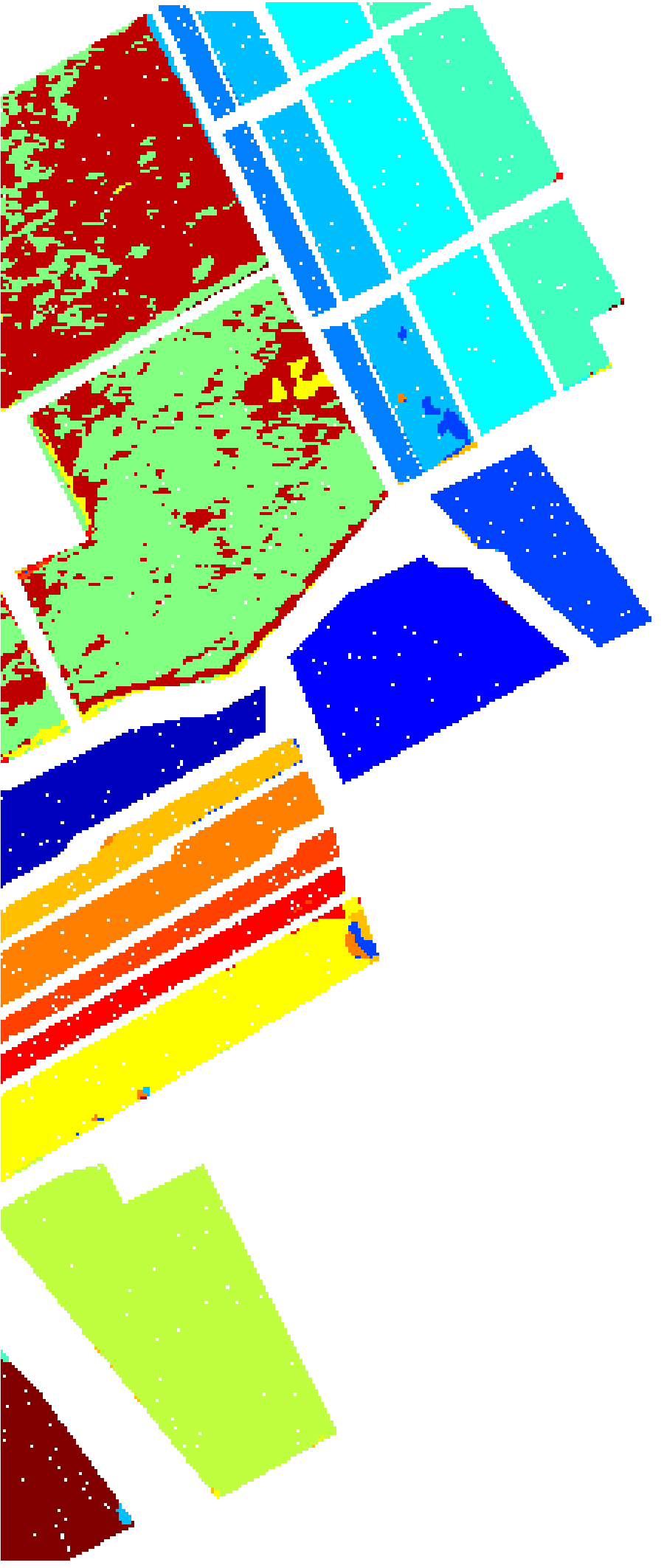}}}\hspace{0pt}	
		
		\subfigure[]{%
			\resizebox*{2.2cm}{4.4cm}{\includegraphics{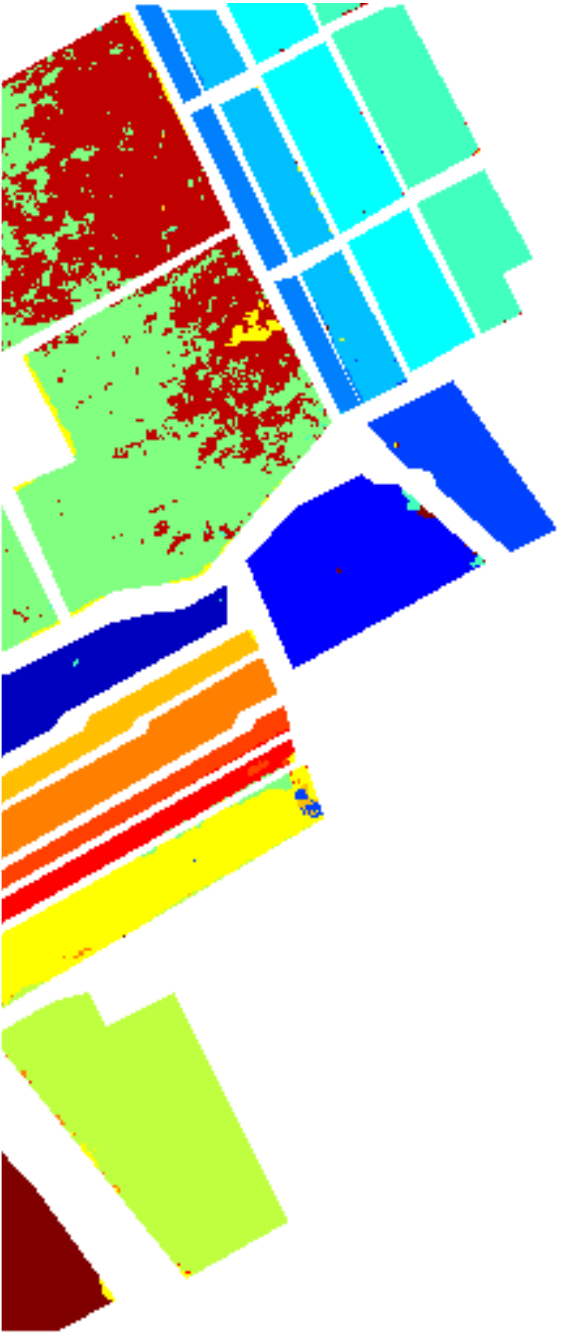}}}\hspace{40pt}	
		\subfigure[]{%
			\resizebox*{2.2cm}{4.4cm}{\includegraphics{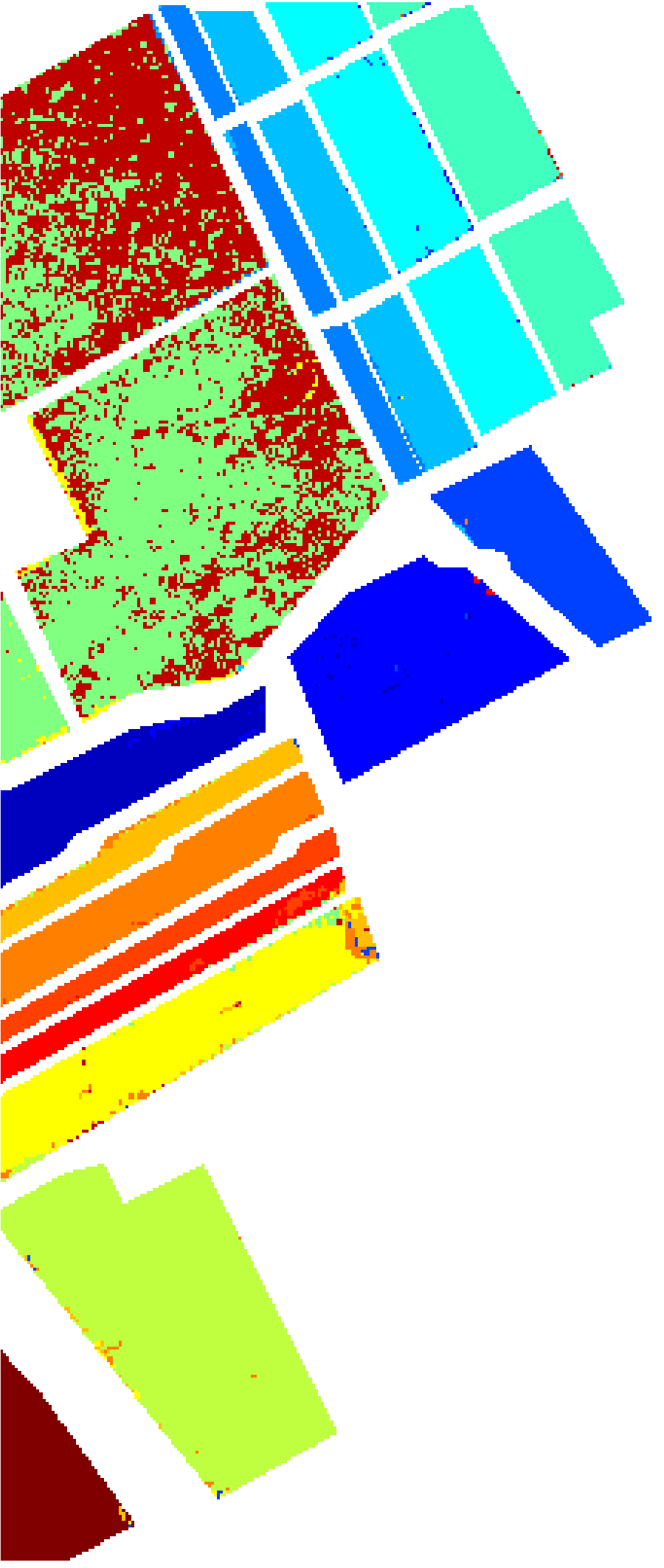}}}\hspace{40pt}
		\subfigure[]{%
			\resizebox*{2.2cm}{4.4cm}{\includegraphics{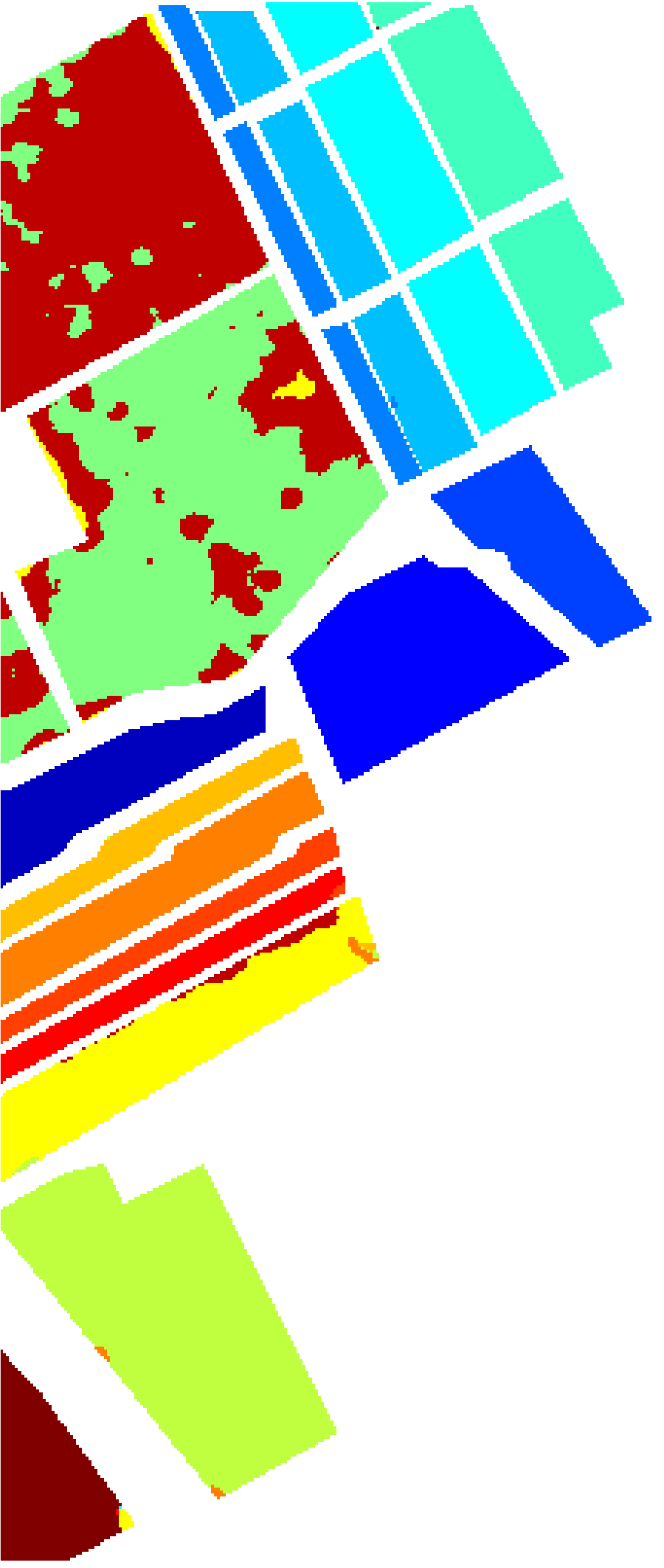}}}\hspace{40pt}			
		\subfigure[]{%
			\label{SAclassificationmap_MGCN_GL}
			\resizebox*{2.2cm}{4.4cm}{\includegraphics{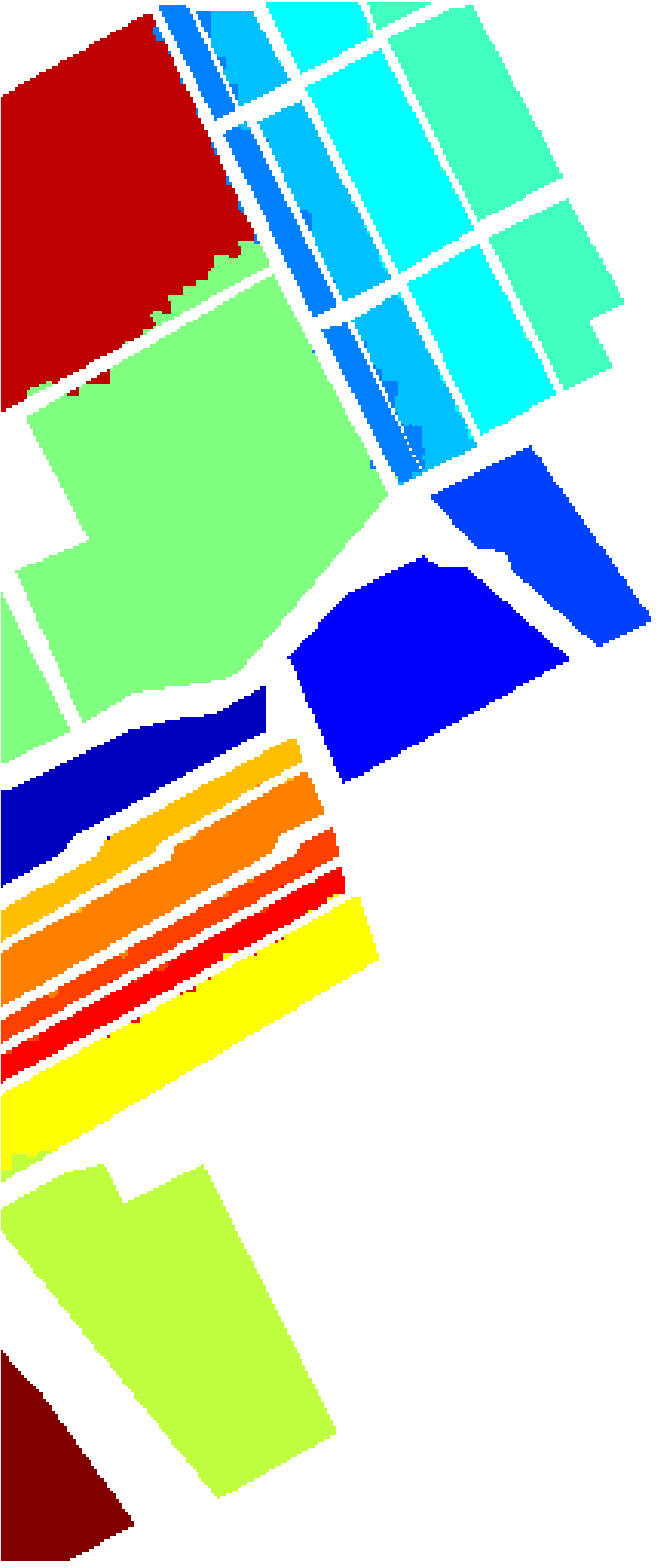}}}\hspace{0pt}
		
		\subfigure {%
			\resizebox*{!}{0.2cm}{\includegraphics{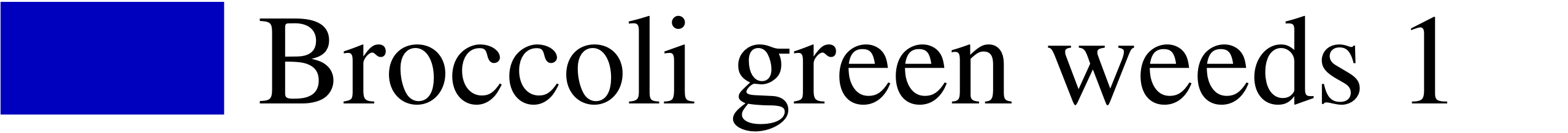}}}\hspace{1pt}
		\subfigure {%
			\resizebox*{!}{0.2cm}{\includegraphics{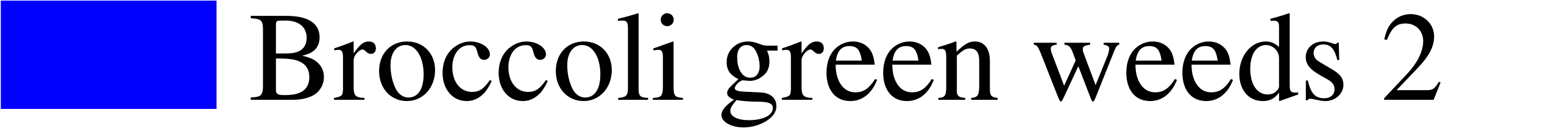}}}\hspace{1pt}
		\subfigure {%
			\resizebox*{!}{0.18cm}{\includegraphics{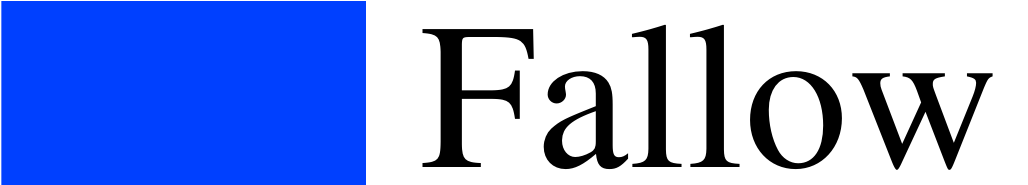}}}\hspace{1pt}
		\subfigure {%
			\resizebox*{!}{0.2cm}{\includegraphics{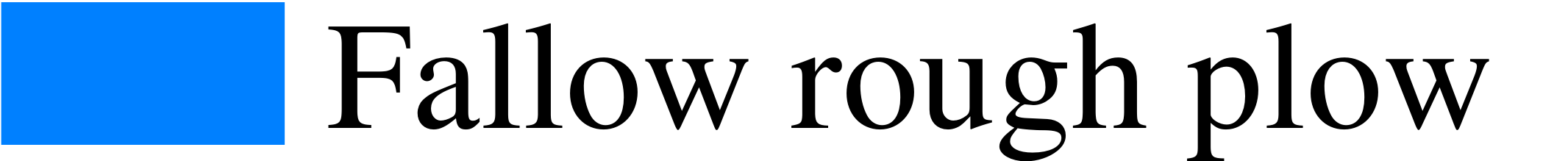}}}\hspace{1pt}
		\subfigure {%
			\resizebox*{!}{0.2cm}{\includegraphics{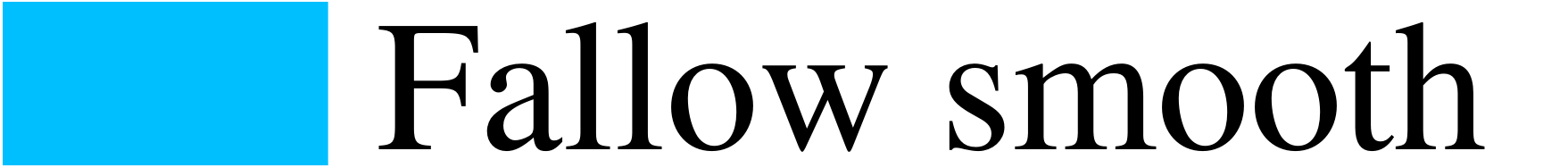}}}\hspace{0pt}
		\subfigure {%
			\resizebox*{!}{0.2cm}{\includegraphics{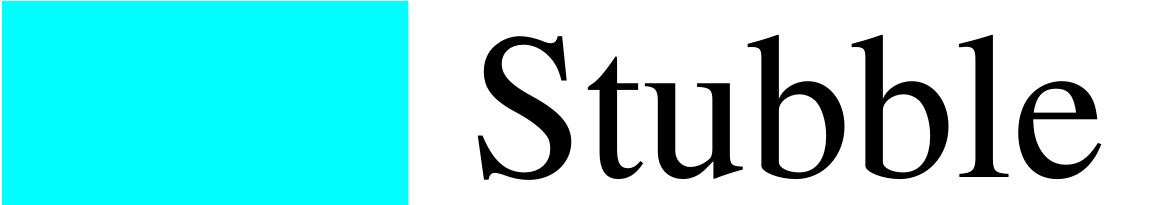}}}\hspace{1pt}	
		\subfigure {%
			\resizebox*{!}{0.205cm}{\includegraphics{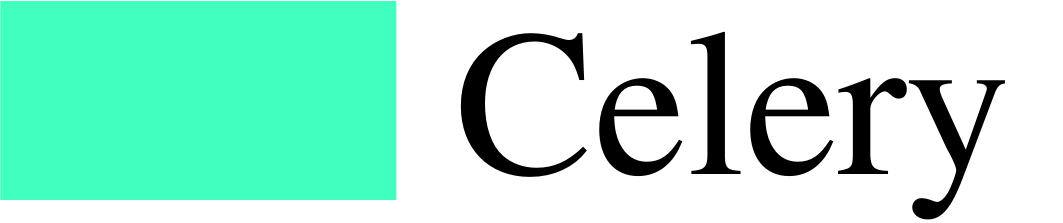}}}\hspace{1pt}
		\subfigure {%
			\resizebox*{!}{0.2cm}{\includegraphics{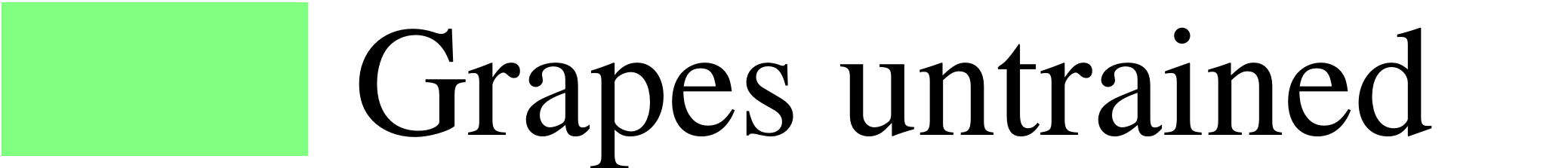}}}\hspace{0pt}
		\subfigure {%
			\resizebox*{!}{0.2cm}{\includegraphics{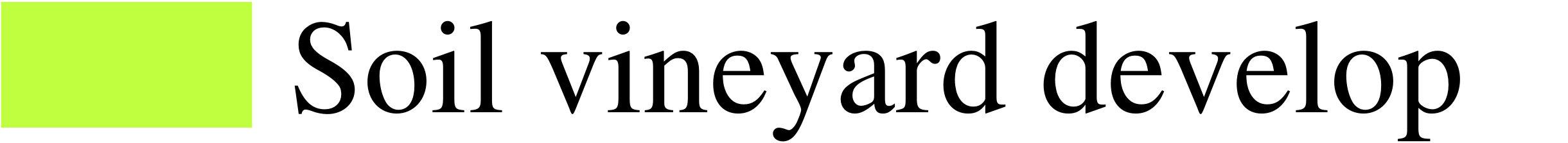}}}\hspace{0pt}	
		\subfigure {%
			\resizebox*{!}{0.2cm}{\includegraphics{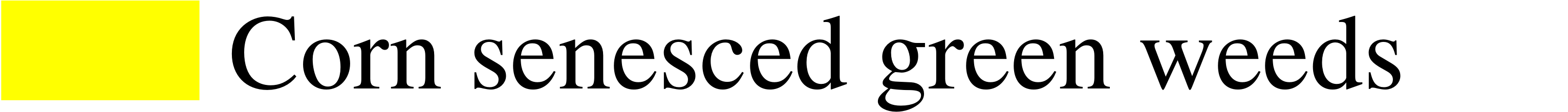}}}\hspace{0pt}
		\subfigure {%
			\resizebox*{!}{0.2cm}{\includegraphics{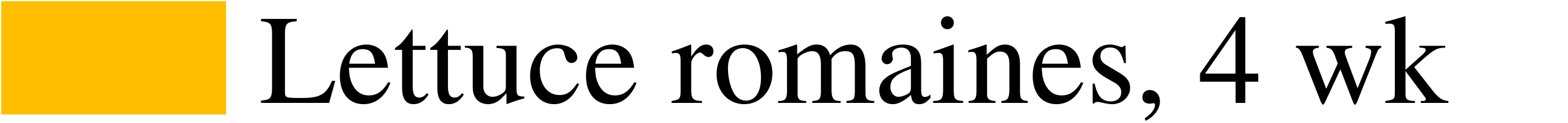}}}\hspace{0pt}
		\subfigure {%
			\resizebox*{!}{0.2cm}{\includegraphics{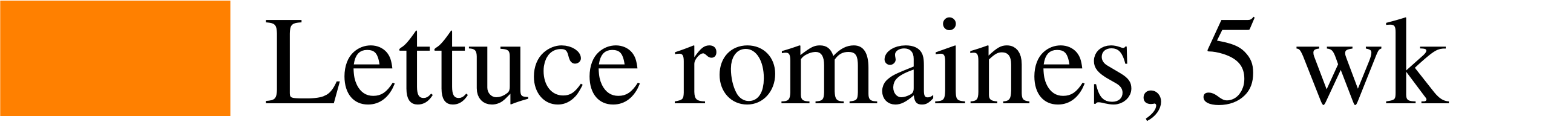}}}\hspace{0pt}		
		\subfigure {%
			\resizebox*{!}{0.2cm}{\includegraphics{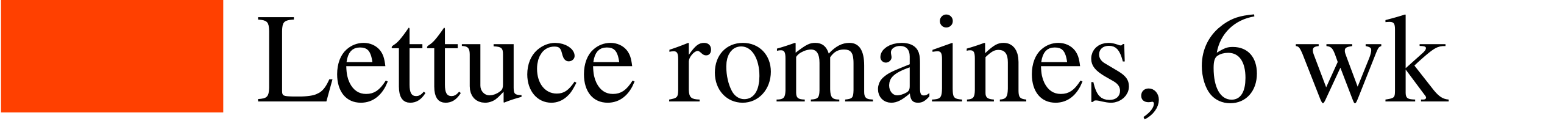}}}\hspace{0pt}	
		\subfigure {%
			\resizebox*{!}{0.2cm}{\includegraphics{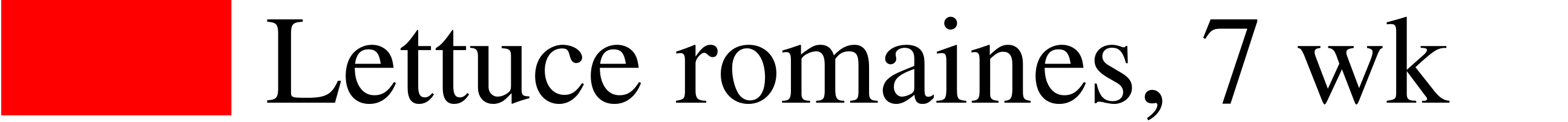}}}\hspace{0pt}
		\subfigure {%
			\resizebox*{!}{0.2cm}{\includegraphics{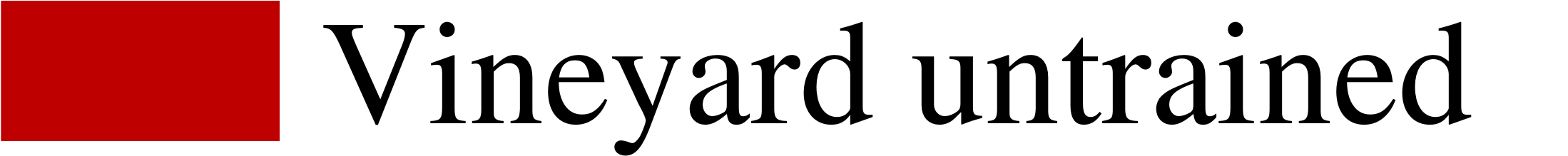}}}\hspace{0pt}		
		\subfigure {%
			\resizebox*{!}{0.2cm}{\includegraphics{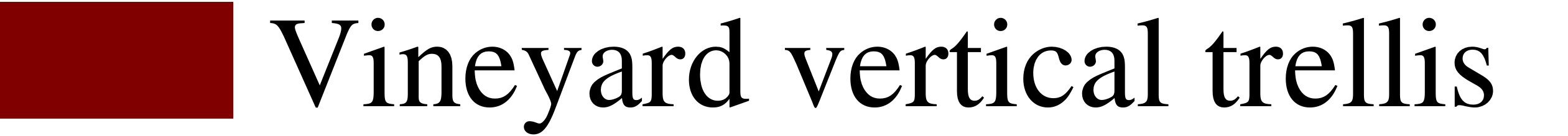}}}\hspace{0pt}	
		\vskip -5pt
		\caption{Classification maps obtained by different methods on Salinas dataset. (a) Ground truth map; (b) S$^{2}$GCN; (c) MDGCN; (d) DR-CNN; (e) CNN-PPF; (f) MFL; (g) JSDF; (h) MGCN-AGL.}  
		\label{SAClassificationMaps}
	\end{figure*}
	
	\subsubsection{Results on the Salinas Dataset}
	
	In Table~\ref{SAClassificationResults}, we quantitatively evaluate the classification performance of different methods on the Salinas dataset. Here, C1-C16 represent the Broccoli green weeds 1, Broccoli green weeds 2, Fallow, Fallow rough plow, Fallow smooth, Stubble, Celery, Grapes untrained, Soil vineyard develop, Corn senesced green weeds, Lettuce romaines, 4 wk, Lettuce romaines, 5 wk, Lettuce romaines, 6 wk, Lettuce romaines, 7 wk, Vineyard untrained, and Vineyard vertical trellis, respectively. Clearly, the performance of our MGCN-AGL is better than that of the baseline methods, especially in C8 (Grapes untrained) and C15 (Vineyard untrained). Meanwhile, the promising results achieved by MDGCN and our MGCN-AGL validate the effectiveness of multi-level spatial information, which has been also proven in the Indian Pines dataset.
	
	A visual comparison can be found in Fig.~\ref{SAClassificationMaps}. We can observe that some pixels in C8 (Grapes untrained) are misclassified into C15 (Vineyard untrained), from which it can inferred that these two land-cover classes have very similar spectral signatures and are a bit difficult to distinguish. However, in Fig.~\ref{SAclassificationmap_MGCN_GL}, our proposed MGCN-AGL still yields smoother visual effect than all the competitors in these two classes.
	
	\begin{figure*}[!t]
		\centering
		\subfigure[]{
			\label{IPclassnum}
			\resizebox*{5cm}{!}{\includegraphics{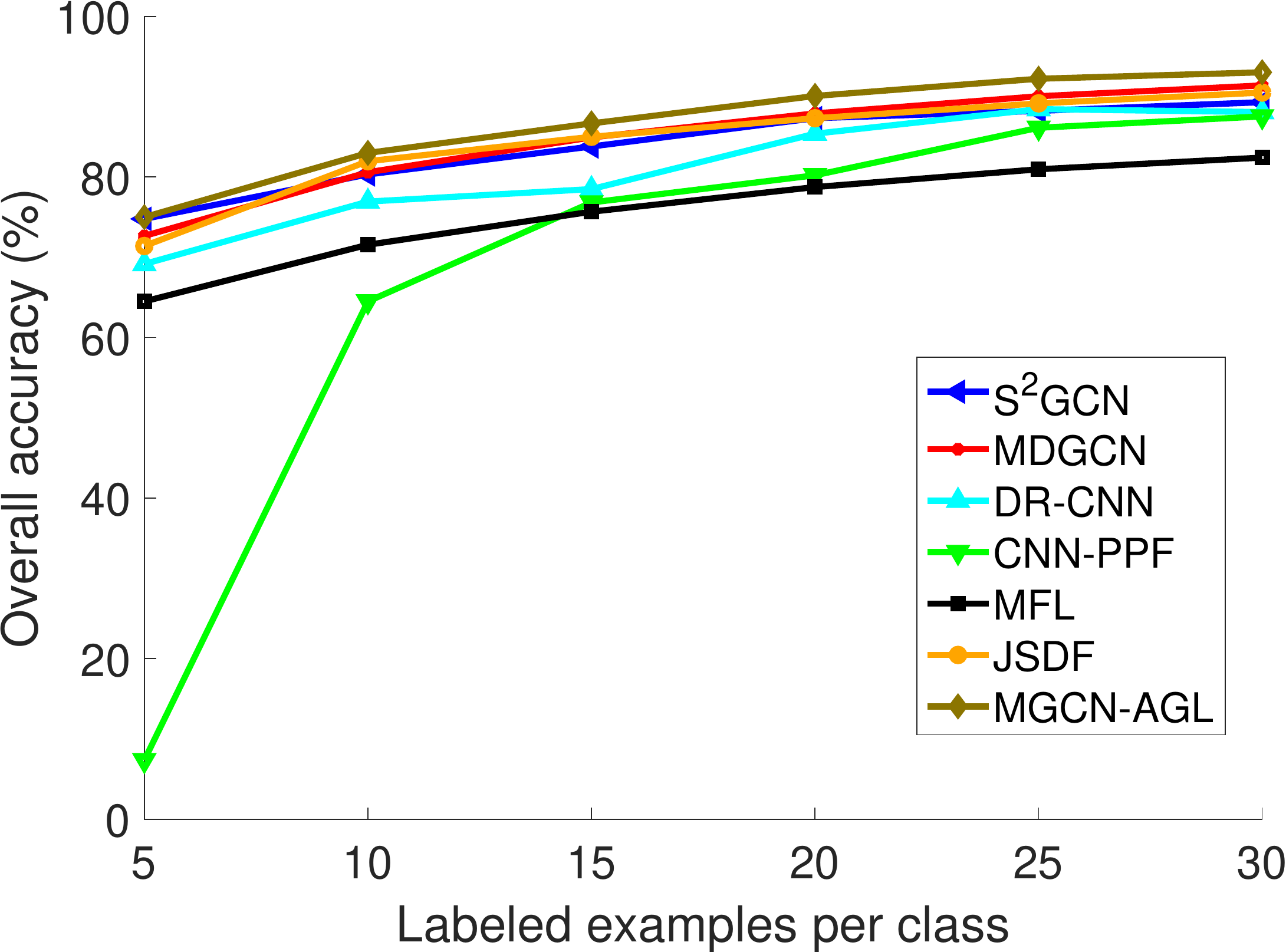}}}\hspace{10pt}
		\subfigure[]{
			\label{PUSclassnum}
			\resizebox*{5 cm}{!}{\includegraphics{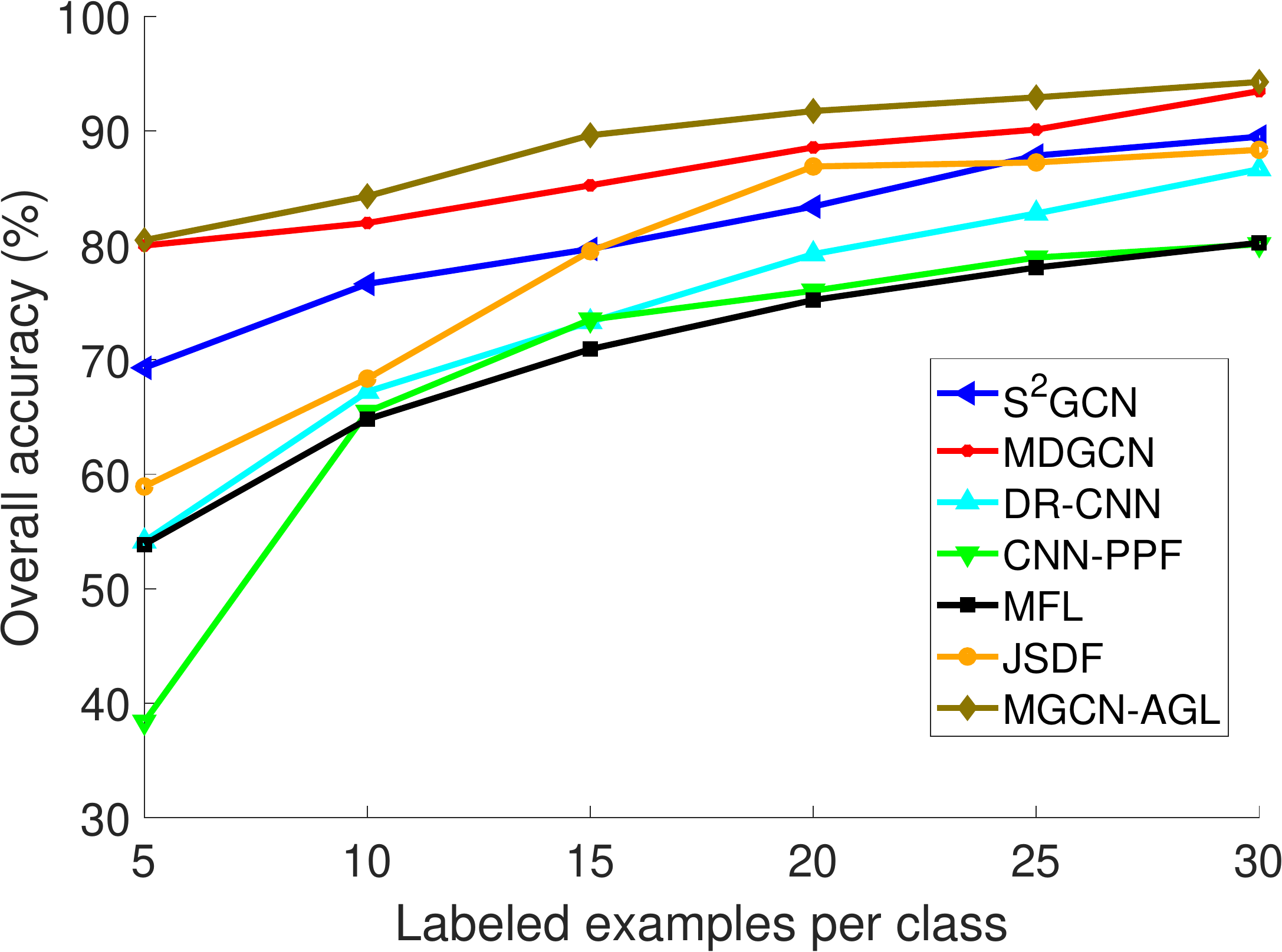}}}\hspace{10pt}
		\subfigure[]{
			\label{SAclassnum}
			\resizebox*{5cm}{!}{\includegraphics{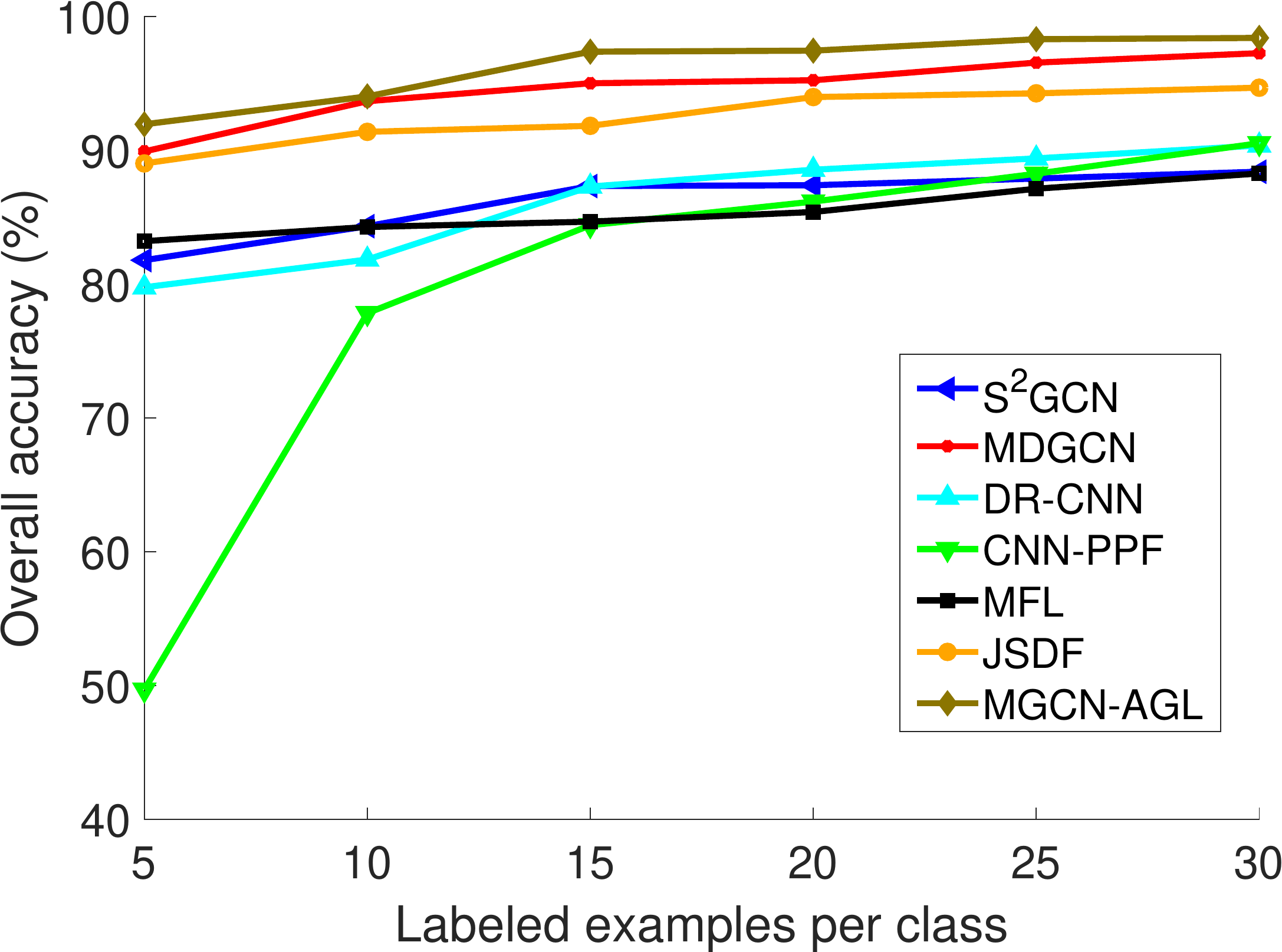}}}
		\vskip -5pt
		\caption{Overall accuracies of various methods under different numbers of labeled examples per class. (a) Houston University dataset; (b) Indian Pines dataset; (c) Salinas dataset.} 
		\label{classnum_acc}
	\end{figure*}
	
	\subsection{Impact of the Number of Labeled Examples}
	
	Fig.~\ref{classnum_acc} presents the classification performance of our proposed MGCN-AGL and the baseline methods under different numbers of initially labeled examples. Specifically, we vary the number of labeled examples per class from 5 to 30 with an interval of 5 and report the OA obtained by all the seven methods on three datasets, i.e., the Houston University, the Indian Pines, and the Salinas. We can observe clearly from these results that MDGCN and our MGCN-AGL generally perform better than the other methods, which illustrates the importance of multi-level spatial information in HSI classification. Meanwhile, our proposed MGCN-AGL allows to learn improved graph information in an automatic manner, which is more robust than using a pre-computed fixed graph. Therefore, the proposed MGCN-AGL is able to achieve the best performance among all the methods. It is also worth mentioning that the performance of the proposed MGCN-AGL is relatively stable with the changed numbers of labeled examples. All these observations demonstrate the effectiveness and stability of our MGCN-AGL.
	
	\begin{figure}[!t]
		\centering
		\resizebox*{6.5cm}{!}{\includegraphics{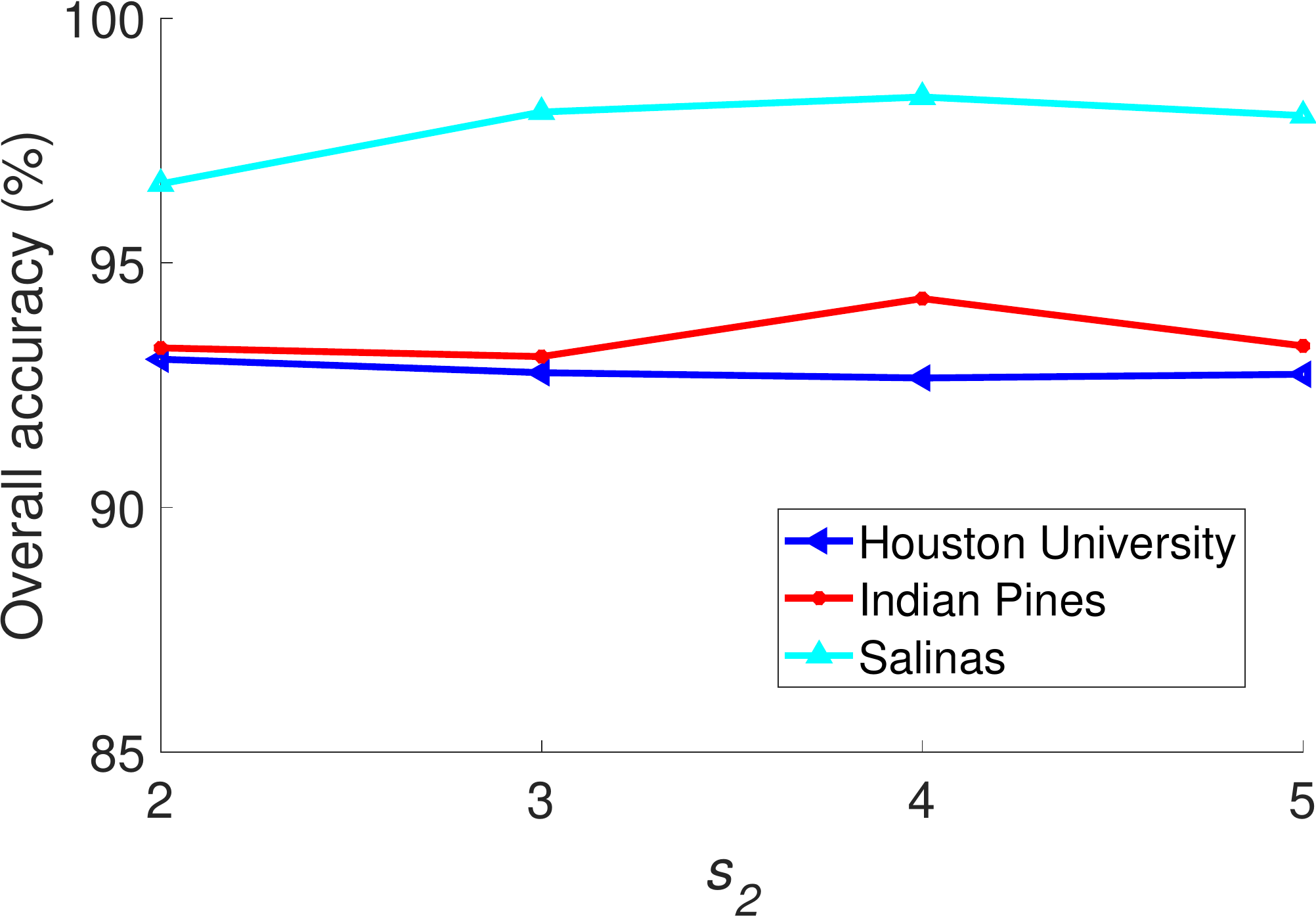}}\hspace{20pt}
		\caption{Parametric sensitivity of the neighborhood size $s_2$ on different datasets.} 
		\label{scale_s_2}
		
	\end{figure}

	\subsection{Parametric Sensitivity}
	\label{para_sen}
	
	In this experiment, we analyze the impact of the neighborhood sizes $s_1$ and $s_2$, which are utilized for incorporating multi-level spatial information, on the classification performance. For parametric simplicity, we fix $s_1$ to 1 and tune the other parameter $s_2$ from 2 to 5. Fig.~\ref{scale_s_2} provides detailed classification results obtained by our method with different values of $s_2$. Observe the curves presented in Fig.~\ref{scale_s_2}, it is remarkable that the selection of neighborhood sizes is critical for our MGCN-AGL to achieve satisfactory performance. For the three hyperspectral datasets, the values of $s_2$ adopted by our method have been shown in Table~\ref{Hyperparameters}.

	\begin{table}[!t]
		\centering
		\caption{OA, AA (\%), and Kappa Coefficient Achieved by Different Model Settings on Houston University Dataset}
		\begin{tabular}{ccc}
			\toprule
			Methods & MGCN-AGL-Loc & MGCN-AGL \\
			\midrule
			OA    & 88.99$\pm$1.64 & \textbf{93.03$\pm$1.02} \\
			AA    & 90.45$\pm$1.51 & \textbf{93.65$\pm$0.94} \\
			Kappa & 88.10$\pm$1.78 & \textbf{92.46$\pm$1.10} \\
			\bottomrule
		\end{tabular}%
		\label{Ablation_UH}%
	\end{table}%
	
	\begin{table}[!t]
		\centering
		\caption{OA, AA (\%), and Kappa Coefficient Achieved by Different Model Settings on Indian Pines Dataset}
		\begin{tabular}{ccc}
			\toprule
			Methods & MGCN-AGL-Loc & MGCN-AGL \\
			\midrule
			OA    & 91.92$\pm$1.66 & \textbf{94.27$\pm$0.92} \\
			AA    & 93.79$\pm$1.07 & \textbf{95.58$\pm$1.18} \\
			Kappa & 90.77$\pm$1.87 & \textbf{93.46$\pm$1.04} \\
			\bottomrule
		\end{tabular}%
		\label{Ablation_IP}%
	\end{table}%
	
	\begin{table}[!t]
		\centering
		\caption{OA, AA (\%), and Kappa Coefficient Achieved by Different Model Settings on Salinas Dataset}
		\begin{tabular}{ccc}
			\toprule
			Methods & MGCN-AGL-Loc & MGCN-AGL \\
			\midrule
			OA    & 96.07$\pm$0.72 & \textbf{98.39$\pm$0.63} \\
			AA    & 96.42$\pm$0.78 & \textbf{98.60$\pm$0.24} \\
			Kappa & 95.63$\pm$0.80 & \textbf{98.21$\pm$0.70} \\
			\bottomrule
		\end{tabular}%
		\label{Ablation_SA}%
	\end{table}%

	\subsection{Ablation Study}
	
	Different from the previous GCN-based HSI classification methods, our proposed MGCN-AGL incorporates the global contextual information to improve the representative ability of the model. Here, we investigate the ablative effect of the global-level convolution. For the sake of comparison, we report the classification result obtained without using the global contextual information, and the reduced model is denoted as `MGCN-AGL-Loc' for simplicity. The number of labeled pixels per class is kept identical to the above experiments in Section \ref{CLassificationResult}. Table~\ref{Ablation_UH}-\ref{Ablation_SA} exhibit the comparative results on the Houston University, the Indian Pines, and the Salinas dataset, respectively. The statistics indicate that the global-level contextual information among image regions is an important supplement to the local spatial information.

	\section{Conclusion}
	\label{Conclusions}
	
	In this paper, we have presented a multi-level graph convolution model, namely MGCN-AGL, to incorporate the global contextual information within the local convolution operation, where the graph information can be automatically learned by the network. By jointly optimizing the learning of feature representations and graph information, our MGCN-AGL allows these two components to benefit each other, which can enhance the expressive power of the generated representations. A comparison of the experimental results with various state-of-the-art HSI classification methods validates the superiority of our proposed MGCN-AGL.

	\ifCLASSOPTIONcaptionsoff
	\newpage
	\fi

	
	
	%

	\bibliographystyle{IEEEtran}
	\bibliography{IEEEabrv,IEEEexample}


\end{document}